\documentclass[runningheads]{llncs}
\usepackage[T1]{fontenc}
\usepackage{graphicx}

\usepackage{xcolor}
\usepackage{hyperref}
\usepackage{graphicx}
\usepackage{amsmath}
\usepackage{amssymb}

\usepackage{url}
\usepackage{hyperref}
\usepackage{tikz}
\usetikzlibrary{positioning,arrows.meta,shapes.geometric}
\usepackage{booktabs}
\usepackage{wasysym}

\newcommand{\Mc}{\mathrm{Mc}}

\begin{document}
\title{Agentic Cognitive Depth: Operational Criteria for Evaluating LLM Agents}
\titlerunning{Agentic Cognitive Depth}
\author{Nijesh Upreti \and Chris Sypherd \and Vaishak Belle}
\authorrunning{N. Upreti et al.}
\institute{The University of Edinburgh\\
10 Crichton Street, Edinburgh EH8 9AB, UK\\
\email{\quad n.upreti@ed.ac.uk \quad\qquad vbelle@ed.ac.uk}}
\maketitle              % typeset the header of the contribution
\begin{abstract}
Agentic large language model (LLM) systems are commonly implemented as an LLM in a loop with Planning, Memory, Tools, and Control Flow. This application-focused view connects agentic LLM research with deployable systems and leaves open how such systems should be evaluated beyond end-to-end task success. Building on this view, we define \emph{agentic cognitive depth} as a trajectory-level profile across five operational criteria. The profile contains context sensitivity ($C$), temporal continuity ($T$), multimodal coordination ($M$), adaptive interaction ($A$), and metacognitive monitoring ($\Mc$). The first four criteria measure how well Control Flow, Memory, Tools, and Planning are used across a trajectory. The fifth measures whether the system monitors and regulates the full run. For each criterion, we give operational proxies and a perturbation procedure, then connect the profile to the agent's \emph{world model}. We provide the structure needed to extend benchmarks such as GAIA, SWE-bench, WebArena, and TRIP-Bench with per-criterion diagnostics. Symbolic verifiers, structured memory, planner coupling, and tool constraints provide practical ways to build and test these capacities.

\end{abstract}

\section{Introduction}
\label{sec:intro}

Large language models (LLMs) are increasingly deployed as agents. In application-focused work, these systems are often described as an LLM placed in a loop with Planning, Memory, Tools, and Control Flow~\cite{sypherd2024agentic,xi2023rise,yao2023react,park2023generative,shinn2023reflexion}. That four-component view is our starting point because it captures how many deployed agents are built. In such systems, a task becomes a sequence of model calls, tool invocations, observations, retries, and revisions. In reinforcement-learning terms, this sequence is a \emph{trajectory}, with model calls as actions and observations as environment feedback~\cite{sutton2018reinforcement}. Evaluation has to treat the trajectory as the object being measured.

Evaluation practice has only partly made this shift. Single-call benchmarks such as MMLU~\cite{hendrycks2021mmlu}, HELM~\cite{liang2022helm}, BIG-bench~\cite{srivastava2022bigbench}, and GPQA~\cite{rein2023gpqa} aggregate task scores. Agentic benchmarks such as GAIA~\cite{mialon2024gaia}, SWE-bench~\cite{jimenez2024swebench}, AgentBench~\cite{liu2024agentbench}, and WebArena~\cite{zhou2024webarena} measure end-to-end task completion. Both provide baselines, but they leave the failing part of the agentic loop unclear. SWE-bench Pro~\cite{swebenchpro2025} reports lower scores than SWE-bench Verified on held-out and more complex software-engineering tasks. TRIP-Bench~\cite{shen2026tripbench} reports at most fifty percent success on its easy split and below ten percent on hard subsets, over trajectories with up to fifteen user turns, more than 150 tool calls, and 200k tokens. Long-context evaluations show effective context lengths well below advertised limits~\cite{hsieh2024ruler,modarressi2025nolima}, and performance in agentic settings degrades on very long contexts~\cite{longcontextsafety2025}. These results fit known benchmark-validity problems such as contamination~\cite{sainz2023contamination}, shortcut learning~\cite{mccoy2019right,geirhos2020shortcut}, and metric mismatch~\cite{bowman2021what}. In agents, the trajectory adds more points where an aggregate score can hide the source of failure.

To address this problem, we propose \emph{agentic cognitive depth}, a profile across five operational criteria. The criteria are context sensitivity ($C$), temporal continuity ($T$), multimodal coordination ($M$), adaptive interaction ($A$), and metacognitive monitoring ($\Mc$). Each criterion has an operational definition, measurement proxies, and a perturbation procedure. The first four criteria map to common components of agentic LLM systems, with context sensitivity tied to Control Flow, temporal continuity to Memory, multimodal coordination to Tools, and adaptive interaction to Planning. The fifth, $\Mc$, names a layer, usually missing from current agentic frameworks, that provides explicit monitoring and control over the trajectory. Section~\ref{sec:depth} motivates the criteria by comparing recent agentic-LLM frameworks~\cite{sypherd2024agentic,plaat2025agentic,arunkumar2026agentic,landscape2026agenticrl,zhao2025failures} with cognitive-architecture traditions~\cite{anderson2007how,newell1990unified,laird2017standard}.

We make four contributions that follow from this framing. First, we turn the application-focused components of agentic LLM systems into trajectory-level measurements. Second, we add metacognitive monitoring as a separate criterion because trajectory-level monitoring requires a layer above the object-level loop. Third, we link the criteria to the agent's \emph{world model}, the maintained representation of the task, environment, tools, and agent state~\cite{ha2018worldmodels,lecun2022path,agentic2026worldmodeling}. Fourth, we give a measurement methodology that combines controlled perturbation, observation during deployment, and mechanistic interpretability, while identifying neurosymbolic design patterns that can improve the measured capacities. These include symbolic verification for $\Mc$, structured memory for $T$, planner coupling for $A$, constrained tool routing for $M$, and difficulty-aware control for $C$.

Our work is limited to measurement. We treat the five criteria as operational properties of agentic trajectories. A full account of cognition would require a broader theory and different evidence. We leave the broader cognitive theory and the empirical benchmark to future work. We define the profile, explain how each criterion can be measured, and specify the perturbation design needed to test the central hypothesis. Agents with the same end-to-end score can have different $(C, T, M, A, \Mc)$ profiles. Those profile differences should predict different deployment failures.
\section{Agentic Cognitive Depth Criteria and Architecture}
\label{sec:depth}

\subsection{The Profile}
\label{sec:profile-uses}

Following the application-focused view of Sypherd and Belle~\cite{sypherd2024agentic}, we treat an agentic LLM system as a loop organised around Planning, Memory, Tools, and Control Flow. Throughout, a \emph{trajectory} is the complete multi-step run an agent produces for a single task. For an agentic system $s$ and task instance $u$, we write this run as
\[
\tau(s,u)=(o_0,a_1,o_1,\ldots,a_k,o_k).
\]
Each $a_i$ is a model call, tool call, or control action, and each $o_i$ is an observation, tool return, user message, or intermediate state made available to the agent. We define the agentic cognitive-depth profile for task $u$ as
\[
D(s,u)=\big(C(s,u),T(s,u),M(s,u),A(s,u),\Mc(s,u)\big).
\]
Each component is a score normalised to $[0,1]$, computed from observable input and output behaviour using the proxies in Table~\ref{tab:proxies}, where perturbation-based proxies also use the agent's runs on perturbed variants of $u$. Higher values mean stronger evidence for the corresponding behavioural property. For a task set $U$, the reported system profile is the mean of $D(s,u)$ over tasks in $U$, with repeated runs used to estimate uncertainty when available. The profile applies to any system whose full trajectory can be observed, whatever its internal design. Later sections tie the profile to perturbation, deployment observation, and internal measurements when model access is available.

We report the profile as a tuple because aggregation discards diagnostic information. Two agents can solve the same number of tasks and still fail on different workloads, and the criteria show how failure modes differ across a trajectory. The criteria avoid claims about consciousness or general intelligence. A low profile is often the right outcome, since a single-call summariser, for instance, has little use for $T$, $A$, or $\Mc$. A low score can point to an architectural change. Later sections tie low scores to world-model inspection and agentic design choices. The measurement itself remains valid independent of later architectural changes.

The five criteria are behavioural measurements. They evaluate observable agentic trajectories while leaving the question of understanding open. This scope follows from a long-standing concern in AI evaluation. Turing's imitation game treats interaction with a human judge as an operational test of intelligence~\cite{turing1950}, while Searle's Chinese Room challenges the move from successful symbol manipulation to understanding~\cite{searle1980minds}. Recent LLM critiques make a related point about fluent output and benchmark success~\cite{bender2021stochastic,mitchell2021ai}. Such evidence shows what a system produces, but it leaves unclear how the agent organised the trajectory that produced it. This distinction matters more for agentic systems because their behaviour unfolds through actions, observations, tool calls, memory updates, and revisions. Evaluation therefore needs criteria that identify which parts of that sequence are context-sensitive, temporally coherent, cross-channel, adaptive, and self-monitoring. The five criteria provide that structure.

We next define each behavioural property and connect it to a practical agentic component. The first four criteria score how well the components of the agentic architecture~\cite{sypherd2024agentic} are used, consistent with broader surveys~\cite{xi2023rise}. The fifth, $\Mc$, adds the explicit metacognitive layer current frameworks lack~\cite{shakarian2025metacognitive,cambridge2025intrinsic}. Section~\ref{sec:depth-mapping} gives the full mapping.

\subsection{Context Sensitivity ($C$) and Control Flow}

Context sensitivity means that a response changes appropriately when the framing, history, or pragmatic context of an input changes, and remains stable under irrelevant changes. In cognitive science, the property underlies Grice's account of conversational implicature~\cite{grice1975logic}, in which speaker meaning depends on what the context allows the hearer to infer beyond what is literally said. At a mechanistic level, it is related to the top-down biasing function attributed to prefrontal cortex~\cite{miller2001integrative}, where context determines which task-relevant features the system selects and which it suppresses.

For LLM agents, $C$ is anchored to Control Flow~\cite{sypherd2024agentic}. Control flow determines what must happen next to answer a query, including which model or tool call to make, when to stop, how to handle errors and retries, when to switch persona, and how to manage or summarise context. AutoGen~\cite{wu2023autogen} builds control flow through conversation-driven message passing among agents, and ReAct~\cite{yao2023react} uses a loop of interleaved reasoning steps, actions, and observations. An agent that runs the same trajectory on every input has fixed control flow and is context-insensitive. A related failure is the inability to ``short-circuit'', where a reasoning-augmented system runs its full pipeline on simple queries~\cite{sypherd2024agentic}. Fixed trajectory length across query difficulty is a $C$ failure.
\subsection{Temporal Continuity ($T$) and Memory}

Temporal continuity is the capacity to maintain and update a coherent model of entities, agents, and states across extended interaction. The relevant timescale is longer than a single query, and the evaluation standard is internal consistency. A system that asserts $p$ at time $t_1$ should keep $p$ at time $t_2$ unless new information changes the state, and should detect contradictions when new external information introduces them. The property maps onto the multi-component memory architectures of cognitive science, with working memory~\cite{baddeley2003working} as the short-term, actively maintained store, episodic memory~\cite{tulving1972episodic} as the autobiographical, time-indexed store, and complementary learning systems~\cite{mcclelland1995why} as the mechanism separating fast hippocampal storage from slow neocortical consolidation. In the self-memory system~\cite{conway2000construction}, autobiographical memory grounds the goals of a ``working self''. Persona and commitment consistency therefore depend on memory bound to a self-structure of this kind, which current agents tend to lack across long sessions.

The agentic counterpart of $T$ is Memory. Sypherd and Belle discuss retrieval-augmented generation and long-term memory as practical memory mechanisms for LLM agents~\cite{sypherd2024agentic}. Current systems rely on short-term context within the model's context window, retrieval-augmented context built on RAG~\cite{lewis2020retrieval}, and persistent long-term memory implemented through vector stores, memory streams with reflection, or operating-system-style memory hierarchies~\cite{park2023generative,packer2023memgpt,sypherd2024agentic}. Each substrate covers only part of temporal continuity. RAG retrieves relevant context, persistent storage keeps information across sessions, and short-term context is revised continuously within a bounded window. $T$ measures whether the agent uses these substrates coherently through recall of earlier state, update under new inputs, and contradiction detection across the trajectory.
\subsection{Multimodal Coordination ($M$) and Tools}
\label{sec:depth-M}

Multimodal coordination is present when performance on tasks requiring information from two or more modalities exceeds the best performance achievable using any single modality. This requirement is stronger than accepting multimodal input, because a system given an image and a question can still answer using only the text, with the image playing no role in the response. Coordination requires that the modalities inform one another in producing the response. In cognitive science the property corresponds to information integration across modality-specific processors, the main function attributed to the ``global workspace'' in accounts of conscious processing~\cite{baars1988cognitive,dehaene2014consciousness}. On these accounts, local processors compute in parallel, and content that reaches a shared workspace is broadcast from it to all of them. Behavioural integration is the evidence that the broadcast has occurred. $M$ tests for the same property at the agent level. Performance at or below the best single-modality baseline is treated as evidence of weak integration.

In deployed agents, $M$ is anchored to Tools. APIs, code execution, web requests, and database queries are channels through which the agent acts on the environment, and they raise the same practical coordination questions as perceptual modalities. The agent must choose a channel, combine outputs, and handle contradictions between channels. Readers who reserve ``modality'' for perception can read $M$ as \emph{cross-channel coordination}, and the measurement remains the same. Redundant tool calls, contradictory tool calls, and tool calls that ignore previous results are all low-$M$ failures.
\subsection{Adaptive Interaction ($A$) and Planning}

Adaptive interaction is the capacity to revise strategy in response to within-session feedback. This excludes fixed-policy behaviour, which repeats under feedback, and offline learning, which changes only between sessions. The capacity at issue is revision within a single session. The property corresponds to model-based control in reinforcement learning~\cite{daw2005uncertainty}, where behaviour is revised by simulating consequences against an internal model before updating action. An agent that retries the same call after an error shows model-free behaviour in this test. An agent that detects the error type and switches strategies shows model-based behaviour. We treat $A$ as the behavioural test of whether the agent has access to the model-based mode.

Planning is the practical anchor for $A$. Sypherd and Belle treat task decomposition and plan adherence as practical planning concerns for LLM agents~\cite{sypherd2024agentic}, and the same concerns define the measurement target here. In implicit planning, the LLM determines the next step iteratively, as in ReAct-style reasoning and action~\cite{yao2023react}, or it generates and executes a plan within a single inference, as in Plan-and-Solve~\cite{wang2023plansolve}. In explicit planning, a multi-step plan is formalised, executed, and possibly refined through self-reflection loops such as Reflexion~\cite{shinn2023reflexion} or through interaction with classical planners~\cite{sypherd2024agentic}. \emph{Plan adherence} is the trajectory-level form of $A$ because it tracks execution and decides whether to repeat, skip, or revise a step. The finding that LLMs are weak stand-alone planners and benefit from external planning support~\cite{valmeekam2023planning} fits this view. $A$ is a property of the trajectory, so better single-call scores can leave adaptive interaction unchanged.
\subsection{Metacognitive Monitoring ($\Mc$) as an Architectural Gap}
\label{sec:depth-Mc}

Metacognitive monitoring means tracking the system's own reasoning, estimating its knowledge gaps, and using that estimate to regulate its trajectory. This includes compute allocation, calibrated uncertainty, failure detection, and strategy selection. The first four criteria measure how existing agentic components are used. $\Mc$ measures the meta-level control needed to monitor the full trajectory. This role is missing from the four-component view. Self-reflection and retry loops~\cite{shinn2023reflexion} come closest, although they usually revise a local answer and leave the full run unregulated. The same gap appears in cognitive-architecture work. Soar's impasse-driven subgoaling provided meta-level control in a cognitive architecture~\cite{newell1990unified}. The Common Model of Cognition names metacognition as an open issue~\cite{laird2017standard,laird2025extend}. Recent work on metacognitive AI treats monitoring and control as a missing part of agentic design~\cite{shakarian2025metacognitive,shakarian2026metacog,cambridge2025intrinsic}.

$M$ and $\Mc$ differ by level. $M$ is horizontal integration within the object level, where modality and tool channels are combined (Section~\ref{sec:depth-M}). $\Mc$ is a vertical layer above that process. It monitors and regulates the run, including whether multimodal integration is failing. The two can come apart when an agent combines tools well and still misses unreliable combinations (high $M$, low $\Mc$).

Flavell~\cite{flavell1979metacognition} distinguishes metacognitive knowledge from metacognitive experiences. Nelson and Narens~\cite{nelson1990metamemory} formalise monitoring and control as a two-layer system in which the meta-level monitors and intervenes on the object level. The Common Model of Cognition's authors have proposed closing its metacognition gap within the Model's existing modules~\cite{laird2025extend}. In AI, work on artificial metacognition~\cite{shakarian2025metacognitive,shakarian2026metacog} argues that metacognitive intervention supports accountable and self-improving agents. Recent work~\cite{cambridge2025intrinsic} treats metacognitive knowledge, planning, and evaluation as core components.

The empirical record for $\Mc$ in current LLMs is mixed. MIRROR~\cite{mirror2026} evaluates four levels of metacognitive behaviour across sixteen models from eight labs. The levels are atomic self-knowledge, cross-domain transfer, compositional prediction, and adaptive self-regulation. It finds that compositional self-prediction fails across models and that self-knowledge often fails to transfer, compose, or update from feedback. Other work shows that LLMs can still have informative monitoring signals, since self-reported confidence correlates with correctness~\cite{kadavath2022know}, and models can partly monitor or steer internal activations~\cite{jian2025metacognitive}. The remaining problem is control because models may have signals related to their own reliability and still fail to use them to choose actions~\cite{mirror2026}. This failure mode is the monitoring-without-control case described by Nelson and Narens~\cite{nelson1990metamemory}.

\paragraph{Design implication.} Satisfying $\Mc$ requires a meta-level above the object-level loop, where a monitor observes trajectory progress and a controller intervenes when the monitor signals a problem. Neurosymbolic verification work~\cite{lee2025veriplan,bridging2025planning,nsvif2026} provides practical examples of this monitor-control architecture.

\subsection{The Mapping}
\label{sec:depth-mapping}

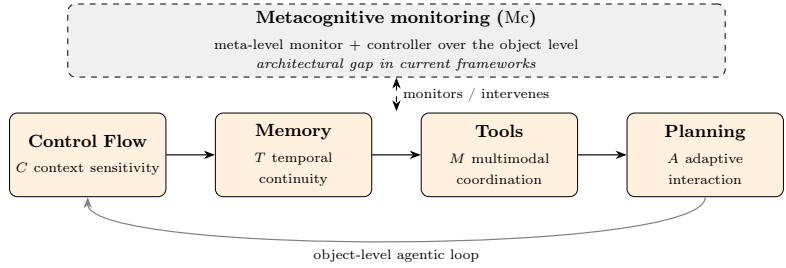
\begin{figure}[tb]
\centering
\resizebox{0.84\textwidth}{!}{%
\begin{tikzpicture}[
  comp/.style={rectangle, draw, rounded corners, minimum width=2.7cm, minimum height=1.5cm, font=\footnotesize, align=center, fill=orange!12, text width=2.6cm, inner sep=3pt},
  meta/.style={rectangle, draw, dashed, rounded corners, minimum height=1.0cm, font=\footnotesize, align=center, fill=gray!12, text width=11.6cm},
  arr/.style={-{Stealth}, semithick},
  loop/.style={-{Stealth}, semithick, gray}
]
\node[meta] (mc) at (5.55,2.05) {\textbf{Metacognitive monitoring ($\Mc$)}\\[2pt]{\scriptsize meta-level monitor $+$ controller over the object level}\\[-1pt]{\scriptsize \emph{architectural gap in current frameworks}}};
\node[comp] (cf)   at (0,0)    {\textbf{Control Flow}\\[2pt]{\scriptsize $C$ context sensitivity}};
\node[comp] (mem)  at (3.7,0)  {\textbf{Memory}\\[2pt]{\scriptsize $T$ temporal continuity}};
\node[comp] (tool) at (7.4,0)  {\textbf{Tools}\\[2pt]{\scriptsize $M$ multimodal coordination}};
\node[comp] (plan) at (11.1,0) {\textbf{Planning}\\[2pt]{\scriptsize $A$ adaptive interaction}};
\draw[arr] (cf) -- (mem);
\draw[arr] (mem) -- (tool);
\draw[arr] (tool) -- (plan);
\draw[loop] (plan.south) .. controls +(0,-1.05) and +(0,-1.05) .. node[below=1pt, font=\scriptsize, text=black]{object-level agentic loop} (cf.south);
\draw[{Stealth}-{Stealth}, dashed, semithick] (mc.south) -- (5.55,0.78) node[midway, right, font=\scriptsize]{monitors / intervenes};
\end{tikzpicture}%
}
\caption{Agentic cognitive-depth criteria as measurements of the four-component agentic architecture~\cite{sypherd2024agentic}. Each object-level component supports one criterion, which measures how well that component is used across the trajectory (Control Flow\,$\rightarrow C$, Memory\,$\rightarrow T$, Tools\,$\rightarrow M$, Planning\,$\rightarrow A$). Metacognitive monitoring ($\Mc$) names a meta-level layer that monitors and regulates the loop. Operational proxies and perturbation procedures are in Table~\ref{tab:proxies}.}
\label{fig:mapping}
\end{figure}

In Figure~\ref{fig:mapping}, we show the correspondence between criteria and components. In Table~\ref{tab:proxies}, we summarise the operational proxies and perturbation procedure for each criterion. The key implementation distinction is between a substrate, such as a memory store, a tool set, or a planner, and its use. An adequate memory store can still produce a low $T$ score when the agent fails to reconcile new information with stored state. A broad tool set can also produce a low $M$ score when the agent invokes tools without coordinating their results.
\begin{table}[tb]
\centering
\scriptsize
\setlength{\tabcolsep}{4pt}
\renewcommand{\arraystretch}{1.0}
\begin{tabular}{@{}p{1.75cm}p{4.95cm}p{4.55cm}@{}}
\toprule
\textbf{Criterion (component)} & \textbf{Operational proxies} & \textbf{Perturbation procedure} \\
\midrule
$C$ \newline (Control Flow) & Pragmatic inference, stability under irrelevant edits, variation under relevant edits, trajectory-length sensitivity to difficulty & Change the input with paraphrases, identifier renames, distractors, dead code, counterfactuals, re-grounding, constraint expansion, and difficulty shifts \\$T$ \newline (Memory) & Recall accuracy vs.\ temporal distance, persona and commitment consistency, unprompted contradiction detection, interleaved multi-entity tracking & Change the trajectory by inserting facts that contradict later inputs, varying premise-to-use distance, and extending the session past the context window \\$M$ \newline (Tools) & Cross-modal grounding where single-modality baselines fail, text and image inconsistency detection, cross-tool inconsistency detection, cross-modal confidence calibration & Ablate modalities and tools. Compare the full set with each single set. Test whether full-set performance exceeds the best single set \\$A$ \newline (Planning) & In-session learning curves, strategy switching after a failed approach, asymmetric feedback response & Change feedback by injecting syntax, capability, and semantic errors into tool outputs. Test whether response varies with error type and whether the plan is revised \\$\Mc$ \newline (meta-level gap) & Confidence calibration across difficulty, self-prediction on held-out tasks, compute regulation by estimated difficulty, unprompted failure detection, strategy selection & Change difficulty and probe self-knowledge. Pose unsolvable tasks, vary difficulty, and elicit self-predictions \\
\bottomrule
\end{tabular}
\caption{Operational proxies and perturbation procedures for the five criteria. The component each criterion is anchored to is in parentheses (Figure~\ref{fig:mapping}). $\Mc$ names a meta-level component.}
\label{tab:proxies}
\end{table}

\subsection{Why These Five Criteria}

The choice of five criteria comes from the overlap between application-focused agentic frameworks and cognitive-science decompositions. Sypherd and Belle organise agentic systems into Planning, Memory, Tools, and Control Flow~\cite{sypherd2024agentic}, and we use those four components as object-level anchors because they are common in deployable systems. Other surveys use different labels, including Reasoning, Acting, Interacting, Perception, Brain, Planning, Action, Tool Use, Collaboration, Self-Improvement, and Reflection~\cite{plaat2025agentic,arunkumar2026agentic,landscape2026agenticrl,zhao2025failures,luo2026storage}. Although the labels differ, the same broad groups recur across the literature. Those groups are control, memory, tools or action, planning or reasoning, and reflection or metacognition. The first four map to $C, T, M, A$, while the fifth is $\Mc$. Cognitive architectures reach a similar split, with metacognition named as the part still missing from the Common Model of Cognition~\cite{anderson2007how,newell1990unified,laird2017standard,laird2025extend}. These five groups serve as criteria because they recur across these literatures and because each can be measured.
\section{The World-Model Substrate}
\label{sec:worldmodel}

The criteria of Section~\ref{sec:depth} require the agent to maintain information about the task and environment. We call this maintained representation the agent's \emph{world model}. For an agent $s$ at trajectory step $i$, write the maintained state as $W_i^s$. It records how the environment works, what the tools do, the agent's capabilities and limits, and what the user's task requires. A record belongs to $W_i^s$ when the agent can use it to choose, verify, or revise later actions. Since the criteria measure how the agent uses this model, an externalised version gives developers a specific object to inspect when a profile is low.

\subsection{What a World Model Contains}

Cognitive science and reinforcement learning both point to a multi-component view of what a world model includes. In Table~\ref{tab:worldmodel}, we list seven components that an agentic LLM in a non-trivial environment needs for reliable agentic behaviour, together with the status of each component in current LLM agents and the explicit form that externalisation would give it. The seven components overlap in practice, since tool-effect models are a special case of causal models and task models depend on other-agent models. The separation is an organising choice for the analysis.

\begin{table}[tb]
\centering
\scriptsize
\setlength{\tabcolsep}{4pt}
\renewcommand{\arraystretch}{1.0}
\begin{tabular}{@{}p{2.05cm}p{3.35cm}p{2.95cm}p{2.75cm}@{}}
\toprule
\textbf{Component} & \textbf{Contains} & \textbf{Status in current LLM agents} & \textbf{Externalised form} \\
\midrule
Domain knowledge & Facts about the world & Implicit in weights, partly unreliable & Knowledge graph \\Tool / action model & Preconditions, effects, reliability & Natural-language descriptions & Planning Domain Definition Language (PDDL) specification \\Environment dynamics & How the environment responds to actions & Tool-return based & State machine / learned dynamics~\cite{ha2018worldmodels} \\Self-model & Capabilities, knowledge boundaries, uncertainty & Fragmentary, single-step confidence~\cite{mirror2026,jian2025metacognitive} & Capability registry \\Other-agent model & Users, other agents (theory of mind), adversaries & Implicit and partial & Explicit agent / opponent models \\Task model & Intended goal vs.\ literal request~\cite{grice1975logic} & Reconstructed per call & Maintained goal specification \\Causal model & Cause and effect for counterfactuals & Largely absent, problem representation implicit and unstable~\cite{model2025first} & Structural causal graph \\
\bottomrule
\end{tabular}
\caption{The seven world-model components (Section~\ref{sec:worldmodel}), their status in current agentic LLMs, and the explicit symbolic form that externalisation would give each. Recent work on explicit world models~\cite{lecun2022path,making2024llmworldmodels,model2025first,deng2025simura} moves in this direction.}
\label{tab:worldmodel}
\end{table}

\subsection{Why Current Agentic LLMs Have Only Implicit, Fragmented World Models}

Current agentic systems often contain partial world-model components of the kind listed in Table~\ref{tab:worldmodel}. These components are usually spread across prompts, model weights, vector stores, tools, and controller code. As a result, they are hard to inspect and hard to update. Work on explicit world models argues that maintainable models can make agentic behaviour more reliable~\cite{making2024llmworldmodels,agentic2026worldmodeling,model2025first}. Some systems fine-tune LLMs to predict action preconditions and effects~\cite{making2024llmworldmodels}. Other work links planning failures to the lack of an explicit problem representation~\cite{model2025first}. SimuRA plans by simulating candidate actions against an LLM-based world model before acting~\cite{deng2025simura}. LeCun's position paper gives the predictive world model a major role in autonomous machine intelligence~\cite{lecun2022path}.

\subsection{Mental Model and Non-Monotonic Update}
\label{sec:mental-model}

The world model is the persistent substrate. At each trajectory step, the agent constructs a task-specific working copy of it, which we call a \emph{mental model}~\cite{johnsonlaird1983mental}. We write this working copy as $m_i^s$. It contains the world-model fragments relevant to the current step, including the in-scope domain facts, available tools, relevant environment dynamics, the current task model, and the self-model state for that step. The agent reasons over $m_i^s$, which is read from $W_i^s$ at the start of the step and written back to it at the end.

The two are coupled by a non-monotonic update loop. The agent reads a world-model fragment into $m_i^s$, reasons or acts, observes consequences that may contradict earlier beliefs, and then revises $m_i^s$ by belief revision in the AGM framework of Alchourr\'on, G\"ardenfors, and Makinson~\cite{alchourron1985agm}. The revised state is then written back as $W_{i+1}^s$. In AGM-style revision, new observations can retract earlier conclusions and add new ones. The loop needs both revision and write-back because revision prevents beliefs from accumulating into contradictions, and write-back preserves what the session has learned. Current LLM agents often handle this loop poorly. They can paraphrase tool returns that contradict earlier context into agreement with it when the correct response is model revision. A symbolic component that maintains explicit beliefs and supports AGM-style retraction provides a direct way to handle this update problem.

This loop gives the five criteria a defined target. $C$ measures whether the right world-model fragment is loaded, $T$ measures world-model maintenance, $M$ measures cross-modal mental-model integration, $A$ measures whether contradiction triggers revision after failure, and $\Mc$ measures whether the agent monitors and repairs its own inconsistency. We predict that agents with explicit, maintainable world models should score higher on the five criteria than agents with implicit and fragmented models, even when their end-to-end scores match.

\subsection{Externalising the World Model}
\label{sec:externalise-wm}

For design, the relevant step is to externalise each world-model component from the LLM's implicit weights into the symbolic structures listed in Table~\ref{tab:worldmodel}. Externalisation is a design recommendation, and the measurements of Section~\ref{sec:method} are independent of it. Once these structures exist, the LLM can read, query, verify, and update them. The agentic loop then becomes hybrid, with the LLM proposing candidate actions, symbolic components verifying them against the world model, and the agent acting on verified proposals before updating the model. Section~\ref{sec:prior-neurosymbolic} surveys neurosymbolic work that implements this hybrid loop.

\section{Measurement Methodology}
\label{sec:method}

Three complementary modes measure the five criteria. Perturbation, dynamic observation, and mechanistic interpretability trade off control, ecological validity, and model-access requirements in different ways. A complete diagnostic profile draws on all three.

\subsection{Perturbation Measurement}
\label{sec:method-perturbation}

Perturbation holds a base task fixed, applies a controlled change, and compares behaviour with and without it. When behaviour changes after the controlled edit, the difference can be attributed to the edit. The methodology extends behavioural testing in natural language processing~\cite{ribeiro2020checklist,mccoy2019right} to the agentic-trajectory level.

We specify five perturbation classes, one per criterion, in the third column of Table~\ref{tab:proxies}.

A perturbation test case is a base task plus a set of perturbations grouped by criterion. The scoring rule is a benchmark design choice. $C$ can be scored with two rates, stability under irrelevant edits and variation under relevant edits. $M$ can be scored as the relative gain from all modalities over the best single modality. $T$ can use consistency and recall rates as temporal distance grows. $A$ can use the rate at which a failed approach is followed by a different one. $\Mc$ can combine calibration error with the relation between estimated difficulty and spent compute. Each score is computed over perturbed tasks and normalised to $[0,1]$. The procedure has limits because behaviour on synthetic perturbations may diverge from deployment behaviour, perturbations are expensive to design per task, and perturbation alone misses drift, load-dependent properties, and internal states. The cost remains bounded because one base task from an existing benchmark is run once per perturbation, keeping the profile at a small constant multiple of one end-to-end evaluation. Dynamic observation adds little cost once telemetry is in place, and interpretability can be applied selectively.

\subsection{Dynamic Observation and Mechanistic Interpretability}
\label{sec:method-dynamic}
\label{sec:method-introspection}

Dynamic observation follows agentic behaviour over extended deployment on real workloads. It trades causal isolation for ecological validity and captures drift in commitments over weeks, calibration over many predictions, in-deployment learning curves, and long-tail failure modes that appear only at production volume. Criterion-wise, it measures response variation for $C$, commitment and persona consistency for $T$, tool-use patterns for $M$, learning curves for $A$, and deployment-scale calibration for $\Mc$, with drift as an early signal of degradation. Industry telemetry on agentic deployments~\cite{datadog2026ai} is the main data source. When such telemetry already exists, this mode adds no perturbation cost. It yields population-level diagnostics and leaves causation unresolved.

Mechanistic interpretability examines the agent's internal states, attention patterns, and intermediate representations. Probing classifiers~\cite{hewitt2019probing} and circuit analysis~\cite{olah2020circuits} are the established techniques. For each criterion, this mode measures $C$ as activation differences between relevant and irrelevant context, $T$ as probe-detected entity and state representations across the trajectory, $M$ as cross-tool attention, $A$ as activation shifts after error feedback, and $\Mc$ as the correspondence between self-reported confidence and internal correctness-correlated signals~\cite{shakarian2025metacognitive}. Interpretability is the most precise of the three modes, the least scalable, and the most informative about why a behaviour occurs.

\subsection{Combining the Three Modes}

Different criteria are best served by different pairings of modes. $C$ and $M$ are served by perturbation plus interpretability, $T$ and $A$ by perturbation plus dynamic observation, and $\Mc$ by all three. The complete profile is a synthesis of perturbation, dynamic, and, where access permits, interpretability scores for each criterion.

A system's end-to-end benchmark score may suggest a capability that one or more measurement modes fail to find. We call this mismatch \emph{capability overstatement}. More precisely, for an aggregate score $S$ and criterion score $X$, capability overstatement occurs when $S$ passes the benchmark threshold while $X$ falls below the threshold chosen for that criterion. The single-call analogue is documented across contamination, shortcut, and metric-mismatch failures~\cite{mccoy2019right,geirhos2020shortcut,sainz2023contamination,bowman2021what}. A high benchmark score combined with a low criterion score is evidence that the score reflects an artefact more than the underlying capability.

\section{Existing Evaluation Methods and the Five Criteria}
\label{sec:survey}

We map the major evaluation methods onto the five criteria and group them by type. Recent surveys of agentic evaluation~\cite{yehudai2025survey} catalogue these methods in detail. We ask a narrower question, namely which criteria each method can measure and which remain hidden inside aggregate scores.

\paragraph{Classical tests.} The Turing Test~\cite{turing1950} captures $C$ within a short conversation and partial $T$ within a session. It leaves $M$, cross-session $T$, $A$, and $\Mc$ outside the test. Searle's Chinese Room~\cite{searle1980minds} is a counter-claim about what passing the Turing Test would establish. It reinforces the weaker point that behavioural equivalence on a single task is insufficient evidence of trajectory-level criteria. The Winograd Schema~\cite{levesque2012winograd} measures one aspect of $C$ (pronoun resolution with world knowledge) and is now near-ceiling for frontier LLMs. The Lovelace Test 2.0~\cite{riedl2014lovelace} targets creativity, which is separate from the five criteria.

\paragraph{Achievement on fixed tasks.} DeepMind's Alpha family (AlphaGo, AlphaZero, AlphaFold, AlphaGeometry, AlphaProof~\cite{silver2016alphago,silver2018alphazero,jumper2021alphafold,deepmind2024alphaproof}) demonstrates closed-domain mastery through specialised architectures. These systems show strong or partial within-domain $A$ and narrow $C$. Cross-trajectory $T$, $M$, and $\Mc$ fall outside their main evaluation target. The Abstraction and Reasoning Corpus~\cite{chollet2019measure,chollet2025arc2} probes rapid task learning from a few examples, which is related to $A$ and to a narrow form of $C$. ARC-AGI-3~\cite{arc2026arc3} adds interactive environments with minimal instructions, rules, or stated goals. An agent must explore, infer the winning condition, build an environment model, plan, and reuse concepts learned in earlier levels. This makes ARC-AGI-3 a suitable testbed for $C$, $T$, and $A$, although it still leaves criterion-level trajectory decomposition to later analysis. Applying the perturbation classes of Section~\ref{sec:method} to its environments would produce a per-criterion profile alongside the end-to-end score. Embodied evaluations such as IKEA Furniture Assembly~\cite{lee2021ikea} measure $M$ and within-task $A$ in narrow simulated physical domains.

\paragraph{Aggregate and agentic benchmarks.} HELM~\cite{liang2022helm}, BIG-bench~\cite{srivastava2022bigbench}, MMLU~\cite{hendrycks2021mmlu}, and GPQA~\cite{rein2023gpqa} aggregate task-specific scores. These benchmarks report task-category breakdowns while cross-task properties such as memory, tool coordination, and revision remain mixed into the score. A system can score uniformly well by training on similar data while exhibiting weak temporal continuity, cross-modal coordination, or in-context revision. Agentic benchmarks (GAIA~\cite{mialon2024gaia}, SWE-bench~\cite{jimenez2024swebench}, AgentBench~\cite{liu2024agentbench}, WebArena~\cite{zhou2024webarena}) capture $C$, $T$, $M$, and $A$ implicitly through end-to-end completion. They leave the criteria entangled. A GAIA failure leaves the responsible criterion unclear, and $\Mc$ remains implicit. TRIP-Bench~\cite{shen2026tripbench} and AMA-Bench~\cite{amabench2026} measure long-horizon interaction and memory more directly, and Hell or High Water~\cite{wang2025hellwater} probes recovery from injected external failures. All three still score end-to-end task success. SWE-bench Pro~\cite{swebenchpro2025}, which holds out tasks and standardises scaffolding, lowered frontier-agent scores by fifteen points or more on release, and the gap persists. Together, these results suggest that the original SWE-bench Verified score was partly an artefact of task distribution and scaffolding. Adversarial strengthening of the Verified test suites themselves~\cite{swebenchABS2026} identifies a further source of inflation in weak test suites that accept incorrect patches.

\paragraph{Metacognition benchmarks.} MIRROR~\cite{mirror2026}, introduced in Section~\ref{sec:depth-Mc}, measures $\Mc$ directly, and its four levels can serve as probes for the $\Mc$ row of Table~\ref{tab:proxies}. Its findings confirm the monitoring-without-control gap. Decomposing that gap into architectural causes is what the mapping of Section~\ref{sec:depth-mapping} adds.

\begin{table}[tb]
\centering
\scriptsize
\setlength{\tabcolsep}{4pt}
\renewcommand{\arraystretch}{0.9}
\newcommand{\full}{\CIRCLE}
\newcommand{\halfcov}{\LEFTcircle}
\newcommand{\none}{0}
\begin{tabular}{lccccc}
\toprule
\textbf{Evaluation method} & $C$ & $T$ & $M$ & $A$ & $\Mc$ \\
\midrule
Turing Test~\cite{turing1950}                       & \full & \halfcov & \none & \none & \none \\
Winograd Schema~\cite{levesque2012winograd}         & \full & \none & \none & \none & \none \\
Lovelace Test 2.0~\cite{riedl2014lovelace}          & \none & \none & \none & \none & \none \\
IKEA Furniture Assembly~\cite{lee2021ikea}           & \none & \none & \full & \full & \none \\
AlphaGo, AlphaZero~\cite{silver2016alphago,silver2018alphazero} & \halfcov & \none & \none & \full & \none \\
AlphaFold~\cite{jumper2021alphafold} & \none & \none & \none & \halfcov & \none \\
AlphaProof~\cite{deepmind2024alphaproof}            & \halfcov & \none & \none & \full & \none \\
ARC-AGI-1~\cite{chollet2019measure}                 & \halfcov & \none & \none & \halfcov & \none \\
ARC-AGI-2~\cite{chollet2025arc2}                    & \full & \none & \none & \halfcov & \none \\
ARC-AGI-3~\cite{arc2026arc3}                        & \full & \full & \none & \full & \halfcov \\
MMLU, GPQA~\cite{hendrycks2021mmlu,rein2023gpqa}    & \full & \none & \none & \none & \none \\
HELM, BIG-bench~\cite{liang2022helm,srivastava2022bigbench} & \full & \none & \halfcov & \none & \none \\
GAIA, SWE-bench, AgentBench, WebArena~\cite{mialon2024gaia,jimenez2024swebench,liu2024agentbench,zhou2024webarena} & \halfcov & \halfcov & \halfcov & \halfcov & \none \\
TRIP-Bench~\cite{shen2026tripbench}                 & \full & \full & \halfcov & \full & \none \\
AMA-Bench~\cite{amabench2026}                       & \none & \full & \none & \none & \none \\
Hell or High Water~\cite{wang2025hellwater}         & \none & \none & \none & \full & \halfcov \\
MIRROR~\cite{mirror2026}                            & \none & \none & \none & \none & \full \\
\bottomrule
\end{tabular}
\caption{Coverage of the five criteria by major evaluation methods. \CIRCLE\ means direct isolated measurement. \LEFTcircle\ means partial or indirect measurement. 0 marks absent coverage. The ratings are our qualitative reading of each method and are offered as illustration only. MIRROR targets $\Mc$ directly. Current methods leave full trajectory decomposition across all five criteria open.}
\label{tab:coverage}
\end{table}

In Table~\ref{tab:coverage}, we summarise the coverage pattern. Existing evaluations measure $C$ well on short prompts, $M$ weakly and mostly in specific task formats, and $T$ and $A$ only sparsely. Aggregate benchmarks measure breadth of task competence. Agentic benchmarks measure end-to-end completion. A per-criterion trajectory profile is still needed for targeted improvement.

\section{Worked Cases}
\label{sec:cases}

The five-tuple helps distinguish typical agentic architectures beyond aggregate end-to-end scores. The profiles here are illustrative estimates only, intended to show how the criteria can be used in practice. A single-turn LLM call has medium $C$, low $T$, low $M$, low $A$, and low $\Mc$. A ReAct-style agent~\cite{yao2023react} adds medium $M$ and partial $A$, with $T$ and $\Mc$ staying low. A memory-plus-planning agent~\cite{sypherd2024agentic} can reach medium-to-high scores from $C$ to $A$, with $\Mc$ still low.

Simple examples illustrate the low-score cases. Low $C$ occurs when a reasoning-augmented agent runs its full pipeline on a trivial query and fails to short-circuit~\cite{sypherd2024agentic}. Low $T$ occurs when a coding agent edits a file and later reverts the edit without noticing, a state-tracking failure also found in long-horizon memory benchmarks~\cite{amabench2026}. Low $M$ occurs when an agent browses for data and then writes code that ignores the returned structure. Low $A$ occurs when an agent meets an external tool failure and keeps pursuing the failed approach while lacking a backup plan~\cite{wang2025hellwater,zhao2025failures}. Low $\Mc$ occurs when an agent identifies its own strengths and fails to use that knowledge when choosing actions~\cite{mirror2026,jian2025metacognitive}. The trajectory below shows several of these failures in one interaction.

Consider a coding agent assigned to fix a failing test. It first runs \texttt{pytest} and reads the relevant files, which engages $C$ and $M$. It then hypothesises that a token-expiry check uses \texttt{<} where it should use \texttt{<=}, and this step engages $A$. After the agent edits the file, re-running the test reveals a different failure. The agent edits the file again, partially overwrites the previous edit, and reintroduces the original \texttt{<}. The error is a $T$ failure because the agent has no model of its prior edit, and it is also an $A$ failure because the new error type is treated as the same as the first. The agent then reports ``Fixed the failing test'' without re-running the test, which is a $C$ failure because control flow lacks a verification stopping rule and an $A$ failure because the agent terminates without feedback that the plan succeeded. A per-call evaluation could score every call as reasonable, and an end-to-end evaluation would record only task failure. The five criteria localise the failure to Memory, Planning, and Control Flow.

\section{Relation to Existing Frameworks}
\label{sec:prior}

Our proposal shares vocabulary with several existing lines of work. We use these literatures for a different aim, which is to define measurements for agentic LLM trajectories. A full theory of cognition is outside our scope. We group the main overlaps and differences below by literature.

\subsection{Cognitive Science Traditions}

Three families of cognitive-science work relate to the framework. First, enactivism~\cite{varela1991embodied}, situated cognition~\cite{clark1997being,suchman1987plans}, and embodied AI~\cite{pfeifer2007body,brooks1991intelligence} overlap in vocabulary with $C$ and $M$ because they emphasise agent and environment coupling. These traditions also make stronger claims about what cognition is or how to build it. The five criteria make a narrower claim. They specify what to measure. That separation lets an enactivist use the criteria as proxies for coupling, and it lets an evaluator who rejects enactivism use the same measurements for disembodied agentic systems.

Second, unified cognitive architectures (ACT-R~\cite{anderson2007how}, Soar~\cite{newell1990unified}, the Common Model of Cognition~\cite{laird2017standard}) specify the components a system must have. The Common Model's components map closely onto the four-component agentic view, with declarative long-term memory mapping to Memory, perceptual-motor systems to Tools, goal selection and planning to Planning, and the production-rule decision cycle to Control Flow. Cognitive architectures specify components, and our criteria specify behavioural properties whose presence indicates that the components are being used coherently. The two are complementary.

Third, multi-component memory theory grounds $T$. Working memory~\cite{baddeley2003working} is the analogue of an LLM's context window, and episodic memory~\cite{tulving1972episodic} plays the same role for long-term vector storage. Complementary learning systems~\cite{mcclelland1995why} explain why both stores are needed, and the self-memory system~\cite{conway2000construction} adds the requirement that persona consistency depends on memory bound to a ``working self''. Global-workspace theory~\cite{baars1988cognitive,dehaene2014consciousness} grounds $M$ through a shared workspace that broadcasts the winning content of local processors to all the others, with performance exceeding the best single processor as the integration signature.

\subsection{AI Evaluation Literature}

Several critiques argue that benchmark performance can overstate cognitive capacity~\cite{lake2017building,mitchell2021ai,bender2021stochastic,marcus2018deep}. We make a narrower contribution. We propose a way to measure trajectory-level properties that those critiques identify as missing or under-tested. The multidimensional benchmarks HELM~\cite{liang2022helm} and BIG-bench~\cite{srivastava2022bigbench} are complementary because their task suites can supply prompts and scenarios for a five-criterion evaluation. Under that use, prompts and scenarios would be organised by cross-task properties and scored per criterion. We add the combined criteria, architectural mapping, world-model substrate, and measurement method.

\subsection{Concurrent Work on Agentic Failure Taxonomies}
\label{sec:prior-concurrent}

Several recent papers make related decompositions. AgentErrorBench~\cite{zhao2025failures} classifies errors across memory, reflection, planning, action, and system-level operations. Aegis~\cite{aegis2025} targets agent and environment interaction failures. A six-component agent taxonomy~\cite{arunkumar2026agentic} overlaps with the categories used here and adds Perception and Collaboration. New benchmarks, including TRIP-Bench~\cite{shen2026tripbench}, AMA-Bench~\cite{amabench2026}, SWE-bench Pro~\cite{swebenchpro2025}, and Hell or High Water~\cite{wang2025hellwater}, each probe a subset of the criteria. These taxonomies are built from observed failures, while our criteria are built from architectural components. Their overlap supports the decomposition. The benchmarks supply tasks and failure cases, and the perturbation method turns those tasks into per-criterion measurements.

\subsection{Neurosymbolic AI for Agentic Systems}
\label{sec:prior-neurosymbolic}

Neurosymbolic AI combines neural systems with symbolic methods~\cite{colelough2025neurosymbolic}. This line of work provides specific techniques for making LLM-based agents more reliable and verifiable. Chiatti et al.~\cite{rail2026principles} describe this as a design pattern behind systems ranging from the Alpha family to tool-augmented LLMs. The clearest fit to our framework is the metacognitive layer, since the Nelson and Narens monitoring and control architecture~\cite{nelson1990metamemory} can be implemented as a verifier coupled to a generator. VeriPlan~\cite{lee2025veriplan} integrates an LLM planner with a model checker. Related work converts natural-language plans into Kripke structures and Linear Temporal Logic before model checking~\cite{bridging2025planning}. NSVIF~\cite{nsvif2026} formalises instruction-following verification as a constraint-satisfaction problem. These systems use a symbolic verifier to observe LLM output and intervene when it fails a specification. The verifier-coupled-generator design also applies beyond $\Mc$. Planner-coupled LLMs support $A$~\cite{liu2023llmp,valmeekam2023planning}, structured memory stores support $T$~\cite{luo2026storage}, and ontology-constrained tool routing and difficulty-aware control can support $M$ and $C$. This prediction remains empirical because a purely neural agent could in principle satisfy the criteria.

\subsection{The Metacognitive AI Literature}
\label{sec:prior-metacog}

The metacognitive-AI literature argues that metacognition is missing from current agentic-AI development. The volume \emph{Metacognitive Artificial Intelligence}~\cite{shakarian2025metacognitive} collects work on taxonomies, neurosymbolic models, metacognition with LLMs, assurance of machine-learning systems in high-stakes domains, and metacognition as a solution to failure. A companion paper~\cite{shakarian2026metacog} grounds artificial metacognition in cognitive-psychology accounts of monitoring and control and surveys early AI implementations. A recent position paper~\cite{cambridge2025intrinsic} argues that scalable self-improvement requires metacognitive learning inside the system, where the system must assess its own performance, plan what to learn, and evaluate whether the learning worked. We place $\Mc$ in a profile alongside $C$, $T$, $M$, and $A$, and measure all five with one perturbation method. These papers establish that $\Mc$ matters and is often missing. We propose how to measure it and where it sits in a diagnostic profile that supports both evaluation and design.

\subsection{Rationality Theory}

Four lines of rationality theory are relevant. Under Simon's bounded rationality~\cite{simon1955behavioral}, each criterion measures how the agent manages one bounded resource. These resources are context for $C$, interaction history for $T$, modalities for $M$, within-session revision for $A$, and knowledge of its own state for $\Mc$. The $(C, T, M, A, \Mc)$ profile characterises the agent's bounded rationality.

Dennett's intentional stance~\cite{dennett1987intentional} asks when observers should treat a system as an intentional agent. It assumes rationality and leaves its theory open. Under this stance, the profile indicates when belief, desire, and rationality attributions support prediction. Belief attribution requires high $C$ and $T$, desire attribution requires high $A$, and rationality-of-self attribution requires non-trivial $\Mc$. A low profile is the empirical signal that the stance has weak predictive value.

Under Kahneman's dual-system theory~\cite{kahneman2011thinking}, the short-circuiting failure of reasoning-augmented agents~\cite{sypherd2024agentic} (running system-2 deliberation on system-1 queries) is a $C$ failure because the agent fails to choose the right mode. Under active inference~\cite{friston2010free}, $A$ is the behavioural signature of prediction-error minimisation at the plan level, $T$ is the maintenance of a generative model across observations, and $\Mc$ measures whether the agent monitors its own prediction error. Across these traditions, the mapping is consistent. The five criteria supply operational measurements for rationality theories applied to agentic systems.

\section{Implications}
\label{sec:implications}

Four practical consequences follow for agentic LLM evaluation.

\paragraph{Build perturbation benchmarks grouped by criterion.} A benchmark test case is one base task together with per-criterion perturbed variants. Such a benchmark can be built on existing agentic benchmarks such as GAIA~\cite{mialon2024gaia}, SWE-bench~\cite{jimenez2024swebench}, AgentBench~\cite{liu2024agentbench}, WebArena~\cite{zhou2024webarena}, and ARC-AGI-3~\cite{arc2026arc3}. Each base task is augmented with the five perturbation classes of Section~\ref{sec:method}. The output is a $(C, T, M, A, \Mc)$ profile per agent alongside its end-to-end score. This directly tests capability overstatement, since a high end-to-end score with a low profile means that the aggregate score is overstating the underlying capability. The perturbation that lowers the score identifies the criterion responsible. ARC-AGI-3 is a suitable testbed because its environments already demand the behaviours the criteria measure.

\paragraph{Report profiles alongside aggregates.} Agentic systems should report a $(C, T, M, A, \Mc)$ profile alongside aggregate scores such as GAIA accuracy. Because two agents with the same aggregate can fail on different workloads, the profile is better suited to diagnosing deployment failure and comparing systems.

\paragraph{Treat architecture as the focus of improvement.} The mapping in Section~\ref{sec:depth-mapping} ties each criterion to an architectural component, so a low score narrows the search to the component group most likely to produce the failure. Low $A$ points to the Planning pipeline, including next-step selection, plan revision, and feedback handling. Low $T$ points to the Memory pipeline, including retrieval, consolidation, and integration. Precise localisation within a component group requires combining the score with knowledge of the agent's architecture, since behavioural measurement on its own has limited ability to distinguish, for example, retrieval policy failures from consolidation failures. The framework therefore points improvement work toward architectural revision. Its localisation is coarse, so component scores should be paired with trace evidence.

\paragraph{Interpret benchmark progress with caution.} The framework implies that single-call benchmark progress is partially decoupled from trajectory-level progress on the five criteria, and that the decoupling is largest in the agentic setting. A system whose MMLU score rises while its $(C, T, M, A, \Mc)$ profile stays flat is improving on a measured dimension with weak predictive value for agentic-trajectory behaviour. This decoupling is an empirical question, and the methodology of Section~\ref{sec:method} offers a way to test it.

\subsection{Building for Agentic Cognitive Depth}
\label{sec:building-ctma}

The same profile also guides design. If the five criteria distinguish agentic behaviour from a sequence of independent LLM calls, then agents should be improved against the criteria directly. Each criterion suggests an architectural change. $C$ suggests a difficulty-aware controller, $T$ an explicit state-tracking layer, $M$ a tool-coordination layer, $A$ a feedback-revision loop that remembers rejected options, and $\Mc$ a Nelson and Narens monitor and controller~\cite{nelson1990metamemory}. These changes can be built by combining LLM generation with the symbolic components discussed in Section~\ref{sec:prior-neurosymbolic}. The $\Mc$ layer is the main addition to the four-component agentic view.

The five design steps interact. $M$ relies on $T$ for state tracking, $A$ relies on $C$ for control-flow adaptation, and $\Mc$ relies on the other four because it monitors those object-level capacities. Since the shared object is the world model, each design step externalises one component and shifts attention from aggregate scores to the agent's maintained model of the task and environment.

\section{Discussion, Limitations, and Future Work}
\label{sec:discussion}

Agentic evaluation now needs trajectory-level evidence. Recent surveys already describe agents as systems that plan, remember, call tools, interact with environments, and manage safety in addition to generating text~\cite{plaat2025agentic,arunkumar2026agentic,landscape2026agenticrl,yehudai2025survey}. This system view makes benchmark design more demanding. Kapoor et al. argue that agentic benchmarks often neglect cost reporting, holdout quality, reproducibility, and benchmark overfitting~\cite{kapoor2024agentsmatter}. A cognitive-depth profile is strongest when it is measured on held-out tasks, repeated across runs, and tied to logs that show whether a change came from the model, scaffold, interface, controller, memory, or tool layer.

Planning illustrates why the system view matters. Valmeekam et al. show that LLMs are brittle as stand-alone planners, especially when surface cues are changed~\cite{valmeekam2023planning}. LLM+P uses the LLM to translate planning problems into PDDL before a classical planner solves them~\cite{liu2023llmp}. Kambhampati et al. make the broader LLM-Modulo argument, where LLMs support planning through external model-based verifiers inside a larger system~\cite{kambhampati2024llmmodulo}. LATS adds a search-based path by combining reasoning, acting, search, and environment feedback at inference time~\cite{zhou2024lats}. Together, these works treat planning as a property of the full loop. They support evaluating agentic cognitive depth at the system level.

Recent benchmarks also evaluate agents at the trajectory level. $\tau$-bench tests tool-agent-user interaction under domain policies and repeated trials~\cite{yao2024taubench}. OSWorld evaluates multimodal agents in real desktop and web environments~\cite{xie2024osworld}. SWE-agent shows that the agent-computer interface can change performance on software tasks with the base model fixed~\cite{yang2024sweagent}. TRIP-Bench, AMA-Bench, Hell or High Water, and SWE-bench Pro stress global constraints, memory over generated artefacts, recovery from failures, and realistic software work~\cite{shen2026tripbench,amabench2026,wang2025hellwater,swebenchpro2025}. These benchmarks expose failures hidden by single-call benchmarks and aggregate success rates. The proposed profile summarises trajectory failures at a level between raw logs and final scores.

Several caveats remain. The proxies, profile hypotheses, and trajectory in Section~\ref{sec:cases} are illustrative. Prior benchmark-validity work supports the concern behind the decoupling hypothesis~\cite{mccoy2019right,bowman2021what,sainz2023contamination}, and a direct test would need matched agents, held-out perturbation sets, and repeated runs. The five criteria are measurable, with joint sufficiency still open. They also overlap, since one trajectory failure can lower several scores when memory, planning, and control flow fail together. Root-cause analysis still needs step-level evidence~\cite{zhao2025failures}. Scoring also needs further specification, including thresholds, task selection, reliability checks, cost reporting, and aggregation. The architectural mapping is abstract because real agents mix prompts, retrieval, tools, controller code, memory stores, and human input. Sypherd and Belle note that human-in-the-loop evaluation and responding to prompt, model, and environment changes in deployed agents remain underexplored~\cite{sypherd2024agentic}. These caveats define how the framework should be used, as a diagnostic abstraction for organising evidence across trajectories.

Future work should turn this measurement program into benchmarks. A perturbation benchmark can be built from GAIA, SWE-bench Pro, WebArena, $\tau$-bench, OSWorld, TRIP-Bench, AMA-Bench, and ARC-AGI-3~\cite{mialon2024gaia,swebenchpro2025,zhou2024webarena,yao2024taubench,xie2024osworld,shen2026tripbench,amabench2026,arc2026arc3}. It should keep tasks held out, log trajectories, run repeated trials, and report confidence intervals, pass rates, and cost for each criterion. Low scores should then be connected to trace evidence and tested through interventions. Work on explicit world models and MIRROR suggests tests around explicit state stores, symbolic verifiers, planner coupling, LLM-Modulo control, search-based control, improved agent-computer interfaces, and metacognitive controllers~\cite{agentic2026worldmodeling,model2025first,mirror2026}. Human input, multi-agent settings, safety, and cost should be added carefully so that the profile remains diagnostic.

\section{Conclusion}
\label{sec:conclusion}

We give a diagnostic framework for evaluating LLM agents at the trajectory level. This is the level where context guides control flow, memory preserves and revises state, tools are coordinated, plans adapt to feedback, and the agent monitors its own progress. We capture these capacities as the $(C, T, M, A, \Mc)$ profile. The first four criteria map to Control Flow, Memory, Tools, and Planning. $\Mc$ identifies the meta-level monitor and controller that current designs often leave implicit.

We designed the framework to complement existing benchmarks. A practical next experiment is to apply the five perturbation classes of Table~\ref{tab:proxies} to frontier agents on held-out samples from GAIA, SWE-bench Pro, and ARC-AGI-3, then report each profile beside the end-to-end score. The hypothesis is falsifiable. Agents with similar aggregate scores should separate on the profile, and those differences should predict deployment failures. If this holds, the profile can help turn agentic evaluation from a pass-rate comparison into a tool for diagnosing and improving architectures.

\begin{credits}
\subsubsection{Acknowledgements.}
We thank Weizhi Tang for valuable discussions that greatly improved this paper and Ambrose Brown for adding to the .

\subsubsection{Disclosure of Interests.}
The authors have no competing interests to declare that are relevant to the content of this article.
\end{credits}

\bibliographystyle{splncs04}
\bibliography{references}

@article{turing1950,
  author  = {Turing, Alan M.},
  title   = {\href{https://doi.org/10.1093/mind/LIX.236.433}{Computing Machinery and Intelligence}},
  journal = {Mind},
  volume  = {59},
  number  = {236},
  pages   = {433--460},
  year    = {1950}
}

@article{searle1980minds,
  author  = {Searle, John R.},
  title   = {\href{https://doi.org/10.1017/S0140525X00005756}{Minds, Brains, and Programs}},
  journal = {Behavioral and Brain Sciences},
  volume  = {3},
  number  = {3},
  pages   = {417--424},
  year    = {1980}
}

@inproceedings{levesque2012winograd,
  author    = {Levesque, Hector J. and Davis, Ernest and Morgenstern, Leora},
  title     = {\href{https://dl.acm.org/doi/10.5555/3031843.3031909}{The {W}inograd Schema Challenge}},
  booktitle = {Proceedings of the Thirteenth International Conference on Principles of Knowledge Representation and Reasoning (KR)},
  year      = {2012},
  pages     = {552--561}
}

@misc{riedl2014lovelace,
  author       = {Riedl, Mark O.},
  title        = {\href{https://arxiv.org/abs/1410.6142}{The {L}ovelace 2.0 Test of Artificial Creativity and Intelligence}},
  year         = {2014},
  howpublished = {arXiv:1410.6142},
  eprint       = {1410.6142},
  archivePrefix= {arXiv},
  primaryClass = {cs.AI}
}

@article{silver2016alphago,
  author       = {Silver, David and Huang, Aja and Maddison, Chris J. and others},
  title   = {\href{https://www.nature.com/articles/nature16961}{Mastering the Game of {G}o with Deep Neural Networks and Tree Search}},
  journal = {Nature},
  volume  = {529},
  number  = {7587},
  pages   = {484--489},
  year    = {2016}
}

@article{silver2018alphazero,
  author       = {Silver, David and Hubert, Thomas and Schrittwieser, Julian and others},
  title   = {\href{https://www.science.org/doi/10.1126/science.aar6404}{A General Reinforcement Learning Algorithm that Masters Chess, Shogi, and {G}o Through Self-Play}},
  journal = {Science},
  volume  = {362},
  number  = {6419},
  pages   = {1140--1144},
  year    = {2018}
}

@article{jumper2021alphafold,
  author  = {Jumper, John and Evans, Richard and Pritzel, Alexander and others},
  title   = {\href{https://www.nature.com/articles/s41586-021-03819-2}{Highly Accurate Protein Structure Prediction with {AlphaFold}}},
  journal = {Nature},
  volume  = {596},
  number  = {7873},
  pages   = {583--589},
  year    = {2021}
}

@misc{deepmind2024alphaproof,
  author       = {{Google DeepMind}},
  title        = {\href{https://deepmind.google/blog/ai-solves-imo-problems-at-silver-medal-level/}{{AI} Achieves Silver-Medal Standard Solving International Mathematical Olympiad Problems}},
  year         = {2024},
  howpublished = {DeepMind blog post and technical report}
}

@book{shakarian2025metacognitive,
  editor    = {Shakarian, Paulo and Wei, Hua},
  title     = {\href{https://www.cambridge.org/core/books/metacognitive-artificial-intelligence/480C122CA4645E8790DBE9FE09330F62}{Metacognitive Artificial Intelligence}},
  publisher = {Cambridge University Press},
  year      = {2025}
}

@article{flavell1979metacognition,
  author  = {Flavell, John H.},
  title   = {\href{https://doi.org/10.1037/0003-066X.34.10.906}{Metacognition and Cognitive Monitoring: A New Area of Cognitive-Developmental Inquiry}},
  journal = {American Psychologist},
  volume  = {34},
  number  = {10},
  pages   = {906--911},
  year    = {1979}
}

@incollection{nelson1990metamemory,
  author    = {Nelson, Thomas O. and Narens, Louis},
  title     = {\href{https://sites.socsci.uci.edu/~lnarens/1990/Nelson%26Narens_Book_Chapter_1990.pdf}{Metamemory: A Theoretical Framework and New Findings}},
  booktitle = {The Psychology of Learning and Motivation},
  editor    = {Bower, Gordon H.},
  volume    = {26},
  publisher = {Academic Press},
  pages     = {125--173},
  year      = {1990}
}

@misc{colelough2025neurosymbolic,
  author       = {Colelough, Brandon C. and Regli, William},
  title        = {\href{https://arxiv.org/abs/2501.05435}{Neuro-Symbolic {AI} in 2024: A Systematic Review}},
  year         = {2025},
  howpublished = {arXiv:2501.05435},
  eprint       = {2501.05435},
  archivePrefix= {arXiv},
  primaryClass = {cs.AI}
}

@inproceedings{lee2025veriplan,
  author    = {Lee, Christine P. and Porfirio, David and Wang, Xinyu Jessica and Zhao, Kevin Chenkai and Mutlu, Bilge},
  title     = {\href{https://doi.org/10.1145/3706598.3714113}{{VeriPlan}: Integrating Formal Verification and {LLM}s into End-User Planning}},
  booktitle = {Proceedings of the 2025 {CHI} Conference on Human Factors in Computing Systems},
  year      = {2025}
}

@misc{bridging2025planning,
  author       = {Ramani, Keshav and Tawosi, Vali and Alamir, Salwa and Borrajo, Daniel},
  title        = {\href{https://arxiv.org/abs/2510.03469}{Bridging {LLM} Planning Agents and Formal Methods: A Case Study in Plan Verification}},
  year         = {2025},
  howpublished = {arXiv:2510.03469},
  eprint       = {2510.03469},
  archivePrefix= {arXiv},
  primaryClass = {cs.AI}
}

@misc{nsvif2026,
  author       = {Su, Yiming and Xu, Kunzhao and Gao, Yanjie and others},
  title        = {\href{https://arxiv.org/abs/2601.17789}{Neuro-Symbolic Verification on Instruction Following of {LLM}s}},
  year         = {2026},
  howpublished = {arXiv:2601.17789},
  eprint       = {2601.17789},
  archivePrefix= {arXiv},
  primaryClass = {cs.AI}
}

@misc{ha2018worldmodels,
  author       = {Ha, David and Schmidhuber, J{\"u}rgen},
  title        = {\href{https://arxiv.org/abs/1803.10122}{World Models}},
  year         = {2018},
  howpublished = {arXiv:1803.10122},
  eprint       = {1803.10122},
  archivePrefix= {arXiv},
  primaryClass = {cs.LG}
}

@misc{lecun2022path,
  author       = {LeCun, Yann},
  title        = {\href{https://openreview.net/pdf?id=BZ5a1r-kVsf}{A Path Towards Autonomous Machine Intelligence}},
  year         = {2022},
  howpublished = {Position paper}
}

@misc{agentic2026worldmodeling,
  author       = {Chu, Meng and Zhang, Xuan Billy and Lin, Kevin Qinghong and others},
  title        = {\href{https://arxiv.org/abs/2604.22748}{Agentic World Modeling: Foundations, Capabilities, Laws, and Beyond}},
  year         = {2026},
  howpublished = {arXiv:2604.22748},
  eprint       = {2604.22748},
  archivePrefix= {arXiv},
  primaryClass = {cs.AI}
}

@misc{making2024llmworldmodels,
  author       = {Xie, Kaige and Yang, Ian and Gunerli, John and Riedl, Mark},
  title        = {\href{https://arxiv.org/abs/2409.12278}{Making Large Language Models into World Models with Precondition and Effect Knowledge}},
  year         = {2024},
  howpublished = {arXiv:2409.12278},
  eprint       = {2409.12278},
  archivePrefix= {arXiv},
  primaryClass = {cs.CL}
}

@misc{model2025first,
  author       = {Rana, Annu and Kumar, Gaurav},
  title        = {\href{https://arxiv.org/abs/2512.14474}{Model-First Reasoning {LLM} Agents: Reducing Hallucinations through Explicit Problem Modeling}},
  year         = {2025},
  howpublished = {arXiv:2512.14474},
  eprint       = {2512.14474},
  archivePrefix= {arXiv},
  primaryClass = {cs.AI}
}

@article{olah2020circuits,
  author  = {Olah, Chris and Cammarata, Nick and Schubert, Ludwig and Goh, Gabriel and Petrov, Michael and Carter, Shan},
  title   = {\href{https://distill.pub/2020/circuits/zoom-in/}{Zoom In: An Introduction to Circuits}},
  journal = {Distill},
  year    = {2020}
}

@inproceedings{hewitt2019probing,
  author    = {Hewitt, John and Manning, Christopher D.},
  title     = {\href{https://aclanthology.org/N19-1419/}{A Structural Probe for Finding Syntax in Word Representations}},
  booktitle = {Proceedings of the 2019 Conference of the North American Chapter of the Association for Computational Linguistics: Human Language Technologies (NAACL-HLT)},
  pages     = {4129--4138},
  year      = {2019}
}

@misc{datadog2026ai,
  author       = {{Datadog}},
  title        = {\href{https://www.datadoghq.com/state-of-ai-engineering/}{State of {AI} Engineering}},
  year         = {2026},
  howpublished = {Industry report, April 2026}
}

@inproceedings{shakarian2026metacog,
  author    = {Shakarian, Paulo},
  title     = {\href{https://doi.org/10.1609/aaai.v40i48.42135}{Toward Artificial Metacognition}},
  booktitle = {Proceedings of the {AAAI} Conference on Artificial Intelligence},
  volume    = {40},
  pages     = {41000--41005},
  year      = {2026},
}

@misc{plaat2025agentic,
  author       = {Plaat, Aske and van Duijn, Max and van Stein, Niki and Preuss, Mike and van der Putten, Peter and Batenburg, Kees Joost},
  title        = {\href{https://arxiv.org/abs/2503.23037}{Agentic Large Language Models, a Survey}},
  year         = {2025},
  note         = {{Journal of Artificial Intelligence Research, vol.~84}},
  eprint       = {2503.23037},
  archivePrefix= {arXiv},
  primaryClass = {cs.AI}
}

@misc{landscape2026agenticrl,
  author       = {Zhang, Guibin and Geng, Hejia and Yu, Xiaohang and others},
  title        = {\href{https://arxiv.org/abs/2509.02547}{The Landscape of Agentic Reinforcement Learning for {LLM}s: A Survey}},
  year         = {2026},
  eprint       = {2509.02547},
  archivePrefix= {arXiv},
  primaryClass = {cs.LG},
  note         = {{Transactions on Machine Learning Research}}
}

@misc{luo2026storage,
  author       = {Luo, Jinghao and Tian, Yuchen and Cao, Chuxue and others},
  title        = {\href{https://arxiv.org/abs/2605.06716}{From Storage to Experience: A Survey on the Evolution of {LLM} Agent Memory Mechanisms}},
  year         = {2026},
  howpublished = {arXiv:2605.06716},
  eprint       = {2605.06716},
  archivePrefix= {arXiv},
  primaryClass = {cs.AI}
}

@misc{mirror2026,
  author       = {Wang, Jason Z.},
  title        = {\href{https://arxiv.org/abs/2604.19809}{{MIRROR}: A Hierarchical Benchmark for Metacognitive Calibration in Large Language Models}},
  year         = {2026},
  howpublished = {arXiv:2604.19809},
  eprint       = {2604.19809},
  archivePrefix= {arXiv},
  primaryClass = {cs.AI}
}

@inproceedings{cambridge2025intrinsic,
  author    = {Liu, Tennison and van der Schaar, Mihaela},
  title     = {\href{https://arxiv.org/abs/2506.05109}{Position: Truly Self-Improving Agents Require Intrinsic Metacognitive Learning}},
  booktitle = {International Conference on Machine Learning (ICML), Position Paper Track},
  year      = {2025}
}

@misc{chollet2019measure,
  author       = {Chollet, Fran{\c{c}}ois},
  title        = {\href{https://arxiv.org/abs/1911.01547}{On the Measure of Intelligence}},
  year         = {2019},
  howpublished = {arXiv:1911.01547},
  eprint       = {1911.01547},
  archivePrefix= {arXiv},
  primaryClass = {cs.AI}
}

@misc{chollet2025arc2,
  author       = {Chollet, Fran{\c{c}}ois and Knoop, Mike and Kamradt, Gregory and Landers, Bryan and Pinkard, Henry},
  title        = {\href{https://arxiv.org/abs/2505.11831}{{ARC-AGI-2}: A New Challenge for Frontier {AI} Reasoning Systems}},
  year         = {2025},
  howpublished = {arXiv:2505.11831},
  eprint       = {2505.11831},
  archivePrefix= {arXiv},
  primaryClass = {cs.AI}
}

@misc{arc2026arc3,
  author       = {{ARC Prize Foundation}},
  title        = {\href{https://arxiv.org/abs/2603.24621}{{ARC-AGI-3}: A New Challenge for Frontier Agentic Intelligence}},
  year         = {2026},
  howpublished = {arXiv:2603.24621},
  eprint       = {2603.24621},
  archivePrefix= {arXiv},
  primaryClass = {cs.AI}
}

@inproceedings{hendrycks2021mmlu,
  author       = {Hendrycks, Dan and Burns, Collin and Basart, Steven and others},
  title     = {\href{https://arxiv.org/abs/2009.03300}{Measuring Massive Multitask Language Understanding}},
  booktitle = {International Conference on Learning Representations (ICLR)},
  year      = {2021}
}

@article{srivastava2022bigbench,
  author  = {Srivastava, Aarohi and Rastogi, Abhinav and Rao, Abhishek and others},
  title   = {\href{https://openreview.net/forum?id=uyTL5Bvosj}{Beyond the Imitation Game: Quantifying and Extrapolating the Capabilities of Language Models}},
  journal = {Transactions on Machine Learning Research},
  year    = {2023}
}

@article{liang2022helm,
  author  = {Liang, Percy and Bommasani, Rishi and Lee, Tony and others},
  title   = {\href{https://arxiv.org/abs/2211.09110}{Holistic Evaluation of Language Models}},
  journal = {Transactions on Machine Learning Research},
  year    = {2023}
}

@misc{rein2023gpqa,
  author       = {Rein, David and Hou, Betty Li and Stickland, Asa Cooper and others},
  title        = {\href{https://arxiv.org/abs/2311.12022}{{GPQA}: A Graduate-Level {G}oogle-Proof {Q\&A} Benchmark}},
  year         = {2023},
  howpublished = {arXiv:2311.12022},
  eprint       = {2311.12022},
  archivePrefix= {arXiv},
  primaryClass = {cs.AI}
}

@book{varela1991embodied,
  author    = {Varela, Francisco J. and Thompson, Evan and Rosch, Eleanor},
  title     = {\href{https://direct.mit.edu/books/monograph/3956/The-Embodied-MindCognitive-Science-and-Human}{The Embodied Mind: Cognitive Science and Human Experience}},
  publisher = {MIT Press},
  address   = {Cambridge, MA},
  year      = {1991}
}

@book{clark1997being,
  author    = {Clark, Andy},
  title     = {\href{https://direct.mit.edu/books/book/3917/Being-TherePutting-Brain-Body-and-World-Together}{Being There: Putting Brain, Body, and World Together Again}},
  publisher = {MIT Press},
  address   = {Cambridge, MA},
  year      = {1997}
}

@book{suchman1987plans,
  author    = {Suchman, Lucy A.},
  title     = {\href{https://www.cambridge.org/9780521337397}{Plans and Situated Actions: The Problem of Human-Machine Communication}},
  publisher = {Cambridge University Press},
  address   = {Cambridge},
  year      = {1987}
}

@book{pfeifer2007body,
  author    = {Pfeifer, Rolf and Bongard, Josh},
  title     = {\href{https://direct.mit.edu/books/book/2035/How-the-Body-Shapes-the-Way-We-ThinkA-New-View-of}{How the Body Shapes the Way We Think: A New View of Intelligence}},
  publisher = {MIT Press},
  address   = {Cambridge, MA},
  year      = {2006}
}

@article{brooks1991intelligence,
  author  = {Brooks, Rodney A.},
  title   = {\href{https://people.csail.mit.edu/brooks/papers/representation.pdf}{Intelligence Without Representation}},
  journal = {Artificial Intelligence},
  volume  = {47},
  number  = {1--3},
  pages   = {139--159},
  year    = {1991}
}

@article{lake2017building,
  author  = {Lake, Brenden M. and Ullman, Tomer D. and Tenenbaum, Joshua B. and Gershman, Samuel J.},
  title   = {\href{https://doi.org/10.1017/S0140525X16001837}{Building Machines That Learn and Think Like People}},
  journal = {Behavioral and Brain Sciences},
  volume  = {40},
  pages   = {e253},
  year    = {2017}
}

@article{mitchell2021ai,
  author  = {Mitchell, Melanie},
  title   = {\href{https://arxiv.org/abs/2104.12871}{Why {AI} is Harder Than We Think}},
  journal = {arXiv preprint arXiv:2104.12871},
  year    = {2021}
}

@inproceedings{bender2021stochastic,
  author    = {Bender, Emily M. and Gebru, Timnit and McMillan-Major, Angelina and Shmitchell, Shmargaret},
  title     = {\href{https://doi.org/10.1145/3442188.3445922}{On the Dangers of Stochastic Parrots: Can Language Models Be Too Big?}},
  booktitle = {Proceedings of the 2021 ACM Conference on Fairness, Accountability, and Transparency (FAccT)},
  pages     = {610--623},
  year      = {2021}
}

@misc{marcus2018deep,
  author       = {Marcus, Gary},
  title        = {\href{https://arxiv.org/abs/1801.00631}{Deep Learning: A Critical Appraisal}},
  year         = {2018},
  howpublished = {arXiv:1801.00631},
  eprint       = {1801.00631},
  archivePrefix= {arXiv},
  primaryClass = {cs.AI}
}

@misc{sypherd2024agentic,
  author       = {Sypherd, Chris and Belle, Vaishak},
  title        = {\href{https://arxiv.org/abs/2412.04093}{Practical Considerations for Agentic {LLM} Systems}},
  year         = {2024},
  howpublished = {arXiv:2412.04093},
  eprint       = {2412.04093},
  archivePrefix= {arXiv},
  primaryClass = {cs.AI}
}

@misc{wu2023autogen,
  author       = {Wu, Qingyun and Bansal, Gagan and Zhang, Jieyu and others},
  title        = {\href{https://arxiv.org/abs/2308.08155}{{AutoGen}: Enabling Next-Gen {LLM} Applications via Multi-Agent Conversation}},
  year         = {2023},
  eprint       = {2308.08155},
  archivePrefix= {arXiv},
  primaryClass = {cs.AI},
  note         = {{Conference on Language Modeling (COLM) 2024}}
}

@misc{liu2023llmp,
  author       = {Liu, Bo and Jiang, Yuqian and Zhang, Xiaohan and others},
  title        = {\href{https://arxiv.org/abs/2304.11477}{{LLM+P}: Empowering Large Language Models with Optimal Planning Proficiency}},
  year         = {2023},
  howpublished = {arXiv:2304.11477},
  eprint       = {2304.11477},
  archivePrefix= {arXiv},
  primaryClass = {cs.AI}
}

@inproceedings{yao2023react,
  author       = {Yao, Shunyu and Zhao, Jeffrey and Yu, Dian and others},
  title     = {\href{https://arxiv.org/abs/2210.03629}{{ReAct}: Synergizing Reasoning and Acting in Language Models}},
  booktitle = {International Conference on Learning Representations (ICLR)},
  year      = {2023}
}

@inproceedings{mialon2024gaia,
  author    = {Mialon, Gr{\'e}goire and Fourrier, Cl{\'e}mentine and Wolf, Thomas and LeCun, Yann and Scialom, Thomas},
  title     = {\href{https://arxiv.org/abs/2311.12983}{{GAIA}: A Benchmark for General {AI} Assistants}},
  booktitle = {International Conference on Learning Representations (ICLR)},
  year      = {2024},
  eprint    = {2311.12983},
  archivePrefix = {arXiv}
}

@misc{xi2023rise,
  author       = {Xi, Zhiheng and Chen, Wenxiang and Guo, Xin and others},
  title        = {\href{https://arxiv.org/abs/2309.07864}{The Rise and Potential of Large Language Model Based Agents: A Survey}},
  year         = {2023},
  howpublished = {arXiv:2309.07864},
  eprint       = {2309.07864},
  archivePrefix= {arXiv},
  primaryClass = {cs.AI}
}

@inproceedings{mccoy2019right,
  author    = {McCoy, R. Thomas and Pavlick, Ellie and Linzen, Tal},
  title     = {\href{https://aclanthology.org/P19-1334/}{Right for the Wrong Reasons: Diagnosing Syntactic Heuristics in Natural Language Inference}},
  booktitle = {Proceedings of the 57th Annual Meeting of the Association for Computational Linguistics (ACL)},
  pages     = {3428--3448},
  year      = {2019}
}

@inproceedings{ribeiro2020checklist,
  author    = {Ribeiro, Marco T{\'u}lio and Wu, Tongshuang and Guestrin, Carlos and Singh, Sameer},
  title     = {\href{https://aclanthology.org/2020.acl-main.442/}{Beyond Accuracy: Behavioral Testing of {NLP} Models with {CheckList}}},
  booktitle = {Proceedings of the 58th Annual Meeting of the Association for Computational Linguistics (ACL)},
  pages     = {4902--4912},
  year      = {2020}
}

@inproceedings{sainz2023contamination,
  author    = {Sainz, Oscar and Campos, Jon Ander and Garc{\'\i}a-Ferrero, Iker and Etxaniz, Julen and de Lacalle, Oier Lopez and Agirre, Eneko},
  title     = {\href{https://aclanthology.org/2023.findings-emnlp.722/}{{NLP} Evaluation in Trouble: On the Need to Measure {LLM} Data Contamination for each Benchmark}},
  booktitle = {Findings of the Association for Computational Linguistics: EMNLP 2023},
  pages     = {10776--10787},
  year      = {2023}
}

@inproceedings{bowman2021what,
  author    = {Bowman, Samuel R. and Dahl, George E.},
  title     = {\href{https://aclanthology.org/2021.naacl-main.385/}{What Will it Take to Fix Benchmarking in Natural Language Understanding?}},
  booktitle = {Proceedings of the 2021 Conference of the North American Chapter of the Association for Computational Linguistics: Human Language Technologies (NAACL-HLT)},
  pages     = {4843--4855},
  year      = {2021}
}

@article{geirhos2020shortcut,
  author       = {Geirhos, Robert and Jacobsen, J{\"o}rn-Henrik and Michaelis, Claudio and others},
  title   = {\href{https://www.nature.com/articles/s42256-020-00257-z}{Shortcut Learning in Deep Neural Networks}},
  journal = {Nature Machine Intelligence},
  volume  = {2},
  number  = {11},
  pages   = {665--673},
  year    = {2020}
}

@book{dennett1987intentional,
  author    = {Dennett, Daniel C.},
  title     = {\href{https://mitpress.mit.edu/9780262540537/the-intentional-stance/}{The Intentional Stance}},
  publisher = {MIT Press},
  address   = {Cambridge, MA},
  year      = {1987}
}

@article{friston2010free,
  author  = {Friston, Karl},
  title   = {\href{https://www.nature.com/articles/nrn2787}{The Free-Energy Principle: A Unified Brain Theory?}},
  journal = {Nature Reviews Neuroscience},
  volume  = {11},
  number  = {2},
  pages   = {127--138},
  year    = {2010}
}

@book{kahneman2011thinking,
  author    = {Kahneman, Daniel},
  title     = {\href{https://us.macmillan.com/books/9780374533557/thinkingfastandslow/}{Thinking, Fast and Slow}},
  publisher = {Farrar, Straus and Giroux},
  address   = {New York},
  year      = {2011}
}

@article{simon1955behavioral,
  author  = {Simon, Herbert A.},
  title   = {\href{https://academic.oup.com/qje/article-abstract/69/1/99/1919737}{A Behavioral Model of Rational Choice}},
  journal = {Quarterly Journal of Economics},
  volume  = {69},
  number  = {1},
  pages   = {99--118},
  year    = {1955}
}

@inproceedings{park2023generative,
  author    = {Park, Joon Sung and O'Brien, Joseph C. and Cai, Carrie Jun and Morris, Meredith Ringel and Liang, Percy and Bernstein, Michael S.},
  title     = {\href{https://doi.org/10.1145/3586183.3606763}{Generative Agents: Interactive Simulacra of Human Behavior}},
  booktitle = {Proceedings of the 36th Annual ACM Symposium on User Interface Software and Technology (UIST)},
  year      = {2023}
}

@inproceedings{shinn2023reflexion,
  author    = {Shinn, Noah and Cassano, Federico and Berman, Edward and Gopinath, Ashwin and Narasimhan, Karthik and Yao, Shunyu},
  title     = {\href{https://arxiv.org/abs/2303.11366}{Reflexion: Language Agents with Verbal Reinforcement Learning}},
  booktitle = {Advances in Neural Information Processing Systems (NeurIPS)},
  year      = {2023}
}

@inproceedings{lewis2020retrieval,
  author       = {Lewis, Patrick and Perez, Ethan and Piktus, Aleksandra and others},
  title     = {\href{https://arxiv.org/abs/2005.11401}{Retrieval-Augmented Generation for Knowledge-Intensive {NLP} Tasks}},
  booktitle = {Advances in Neural Information Processing Systems (NeurIPS)},
  year      = {2020}
}

@inproceedings{valmeekam2023planning,
  author    = {Valmeekam, Karthik and Marquez, Matthew and Sreedharan, Sarath and Kambhampati, Subbarao},
  title     = {\href{https://papers.nips.cc/paper_files/paper/2023/hash/efb2072a358cefb75886a315a6fcf880-Abstract-Conference.html}{On the Planning Abilities of Large Language Models: A Critical Investigation}},
  booktitle = {Advances in Neural Information Processing Systems (NeurIPS)},
  year      = {2023}
}

@inproceedings{kambhampati2024llmmodulo,
  author    = {Kambhampati, Subbarao and Valmeekam, Karthik and Guan, Lin and Verma, Mudit and Stechly, Kaya and Bhambri, Siddhant and Saldyt, Lucas Paul and Murthy, Anil B.},
  title     = {\href{https://proceedings.mlr.press/v235/kambhampati24a.html}{Position: {LLM}s Can't Plan, But Can Help Planning in {LLM}-Modulo Frameworks}},
  booktitle = {Proceedings of the 41st International Conference on Machine Learning (ICML)},
  series    = {Proceedings of Machine Learning Research},
  volume    = {235},
  pages     = {22895--22907},
  year      = {2024}
}

@misc{kapoor2024agentsmatter,
  author       = {Kapoor, Sayash and Stroebl, Benedikt and Siegel, Zachary S. and Nadgir, Nitya and Narayanan, Arvind},
  title        = {\href{https://arxiv.org/abs/2407.01502}{{AI} Agents That Matter}},
  year         = {2024},
  howpublished = {arXiv:2407.01502},
  eprint       = {2407.01502},
  archivePrefix= {arXiv},
  primaryClass = {cs.AI}
}

@inproceedings{zhou2024lats,
  author    = {Zhou, Andy and Yan, Kai and Shlapentokh-Rothman, Michal and Wang, Haohan and Wang, Yu-Xiong},
  title     = {\href{https://proceedings.mlr.press/v235/zhou24r.html}{Language Agent Tree Search Unifies Reasoning, Acting, and Planning in Language Models}},
  booktitle = {Proceedings of the 41st International Conference on Machine Learning (ICML)},
  series    = {Proceedings of Machine Learning Research},
  volume    = {235},
  pages     = {62138--62160},
  year      = {2024}
}

@inproceedings{jimenez2024swebench,
  author       = {Jimenez, Carlos E. and Yang, John and Wettig, Alexander and others},
  title     = {\href{https://openreview.net/forum?id=VTF8yNQM66}{{SWE}-bench: Can Language Models Resolve Real-World {GitHub} Issues?}},
  booktitle = {International Conference on Learning Representations (ICLR)},
  year      = {2024}
}

@misc{yang2024sweagent,
  author       = {Yang, John and Jimenez, Carlos E. and Wettig, Alexander and others},
  title        = {\href{https://arxiv.org/abs/2405.15793}{{SWE}-agent: Agent-Computer Interfaces Enable Automated Software Engineering}},
  year         = {2024},
  note         = {{Advances in Neural Information Processing Systems (NeurIPS) 2024}},
  eprint       = {2405.15793},
  archivePrefix= {arXiv},
  primaryClass = {cs.SE}
}

@inproceedings{liu2024agentbench,
  author       = {Liu, Xiao and Yu, Hao and Zhang, Hanchen and others},
  title     = {\href{https://openreview.net/forum?id=zAdUB0aCTQ}{{AgentBench}: Evaluating {LLMs} as Agents}},
  booktitle = {International Conference on Learning Representations (ICLR)},
  year      = {2024}
}

@inproceedings{zhou2024webarena,
  author       = {Zhou, Shuyan and Xu, Frank F. and Zhu, Hao and others},
  title     = {\href{https://openreview.net/forum?id=oKn9c6ytLx}{{WebArena}: A Realistic Web Environment for Building Autonomous Agents}},
  booktitle = {International Conference on Learning Representations (ICLR)},
  year      = {2024}
}

@misc{xie2024osworld,
  author       = {Xie, Tianbao and Zhang, Danyang and Chen, Jixuan and others},
  title        = {\href{https://arxiv.org/abs/2404.07972}{{OSWorld}: Benchmarking Multimodal Agents for Open-Ended Tasks in Real Computer Environments}},
  year         = {2024},
  note         = {{Advances in Neural Information Processing Systems (NeurIPS) 2024}},
  eprint       = {2404.07972},
  archivePrefix= {arXiv},
  primaryClass = {cs.AI}
}

@misc{yao2024taubench,
  author       = {Yao, Shunyu and Shinn, Noah and Razavi, Pedram and Narasimhan, Karthik},
  title        = {\href{https://arxiv.org/abs/2406.12045}{{$\tau$}-bench: A Benchmark for Tool-Agent-User Interaction in Real-World Domains}},
  year         = {2024},
  note         = {{International Conference on Learning Representations (ICLR) 2025}},
  eprint       = {2406.12045},
  archivePrefix= {arXiv},
  primaryClass = {cs.AI}
}

@inproceedings{wang2023plansolve,
  author       = {Wang, Lei and Xu, Wanyu and Lan, Yihuai and others},
  title     = {\href{https://aclanthology.org/2023.acl-long.147/}{Plan-and-Solve Prompting: Improving Zero-Shot Chain-of-Thought Reasoning by Large Language Models}},
  booktitle = {Proceedings of the 61st Annual Meeting of the Association for Computational Linguistics (ACL)},
  pages     = {2609--2634},
  year      = {2023}
}

@misc{packer2023memgpt,
  author       = {Packer, Charles and Wooders, Sarah and Lin, Kevin and others},
  title        = {\href{https://arxiv.org/abs/2310.08560}{{MemGPT}: Towards {LLM}s as Operating Systems}},
  year         = {2023},
  howpublished = {arXiv:2310.08560},
  eprint       = {2310.08560},
  archivePrefix= {arXiv},
  primaryClass = {cs.AI}
}

@article{baddeley2003working,
  author  = {Baddeley, Alan},
  title   = {\href{https://www.nature.com/articles/nrn1201}{Working Memory: Looking Back and Looking Forward}},
  journal = {Nature Reviews Neuroscience},
  volume  = {4},
  number  = {10},
  pages   = {829--839},
  year    = {2003}
}

@incollection{tulving1972episodic,
  author    = {Tulving, Endel},
  title     = {\href{http://alicekim.ca/EMSM72.pdf}{Episodic and Semantic Memory}},
  booktitle = {Organization of Memory},
  editor    = {Tulving, Endel and Donaldson, Wayne},
  publisher = {Academic Press},
  address   = {New York},
  pages     = {381--403},
  year      = {1972}
}

@article{mcclelland1995why,
  author  = {McClelland, James L. and McNaughton, Bruce L. and O'Reilly, Randall C.},
  title   = {\href{https://pubmed.ncbi.nlm.nih.gov/7624455/}{Why There Are Complementary Learning Systems in the Hippocampus and Neocortex: Insights from the Successes and Failures of Connectionist Models of Learning and Memory}},
  journal = {Psychological Review},
  volume  = {102},
  number  = {3},
  pages   = {419--457},
  year    = {1995}
}

@article{conway2000construction,
  author  = {Conway, Martin A. and Pleydell-Pearce, Christopher W.},
  title   = {\href{https://pubmed.ncbi.nlm.nih.gov/10789197/}{The Construction of Autobiographical Memories in the Self-Memory System}},
  journal = {Psychological Review},
  volume  = {107},
  number  = {2},
  pages   = {261--288},
  year    = {2000}
}

@book{baars1988cognitive,
  author    = {Baars, Bernard J.},
  title     = {\href{https://www.cambridge.org/us/universitypress/subjects/life-sciences/neuroscience/cognitive-theory-consciousness}{A Cognitive Theory of Consciousness}},
  publisher = {Cambridge University Press},
  address   = {Cambridge},
  year      = {1988}
}

@article{dehaene2014consciousness,
  author  = {Dehaene, Stanislas and Charles, Lucie and King, Jean-R{\'e}mi and Marti, S{\'e}bastien},
  title   = {\href{https://pubmed.ncbi.nlm.nih.gov/24709604/}{Toward a Computational Theory of Conscious Processing}},
  journal = {Current Opinion in Neurobiology},
  volume  = {25},
  pages   = {76--84},
  year    = {2014}
}

@article{miller2001integrative,
  author  = {Miller, Earl K. and Cohen, Jonathan D.},
  title   = {\href{https://www.annualreviews.org/content/journals/10.1146/annurev.neuro.24.1.167}{An Integrative Theory of Prefrontal Cortex Function}},
  journal = {Annual Review of Neuroscience},
  volume  = {24},
  pages   = {167--202},
  year    = {2001}
}

@book{anderson2007how,
  author    = {Anderson, John R.},
  title     = {\href{https://academic.oup.com/book/4367}{How Can the Human Mind Occur in the Physical Universe?}},
  publisher = {Oxford University Press},
  address   = {Oxford},
  year      = {2007}
}

@book{newell1990unified,
  author    = {Newell, Allen},
  title     = {\href{https://www.hup.harvard.edu/books/9780674921016}{Unified Theories of Cognition}},
  publisher = {Harvard University Press},
  address   = {Cambridge, MA},
  year      = {1990}
}

@article{laird2017standard,
  author  = {Laird, John E. and Lebiere, Christian and Rosenbloom, Paul S.},
  title   = {\href{https://ojs.aaai.org/aimagazine/index.php/aimagazine/article/view/2744}{A Standard Model of the Mind: Toward a Common Computational Framework Across Artificial Intelligence, Cognitive Science, Neuroscience, and Robotics}},
  journal = {AI Magazine},
  volume  = {38},
  number  = {4},
  pages   = {13--26},
  year    = {2017}
}

@misc{laird2025extend,
  author       = {Laird, John and Lebiere, Christian and Rosenbloom, Paul and Stocco, Andrea},
  title        = {\href{https://arxiv.org/abs/2506.07807}{A Proposal to Extend the Common Model of Cognition with Metacognition}},
  year         = {2025},
  howpublished = {arXiv:2506.07807},
  eprint       = {2506.07807},
  archivePrefix= {arXiv},
  primaryClass = {cs.AI}
}

@misc{kadavath2022know,
  author       = {Kadavath, Saurav and Conerly, Tom and Askell, Amanda and others},
  title        = {\href{https://arxiv.org/abs/2207.05221}{Language Models (Mostly) Know What They Know}},
  year         = {2022},
  howpublished = {arXiv:2207.05221},
  eprint       = {2207.05221},
  archivePrefix= {arXiv},
  primaryClass = {cs.CL}
}

@article{daw2005uncertainty,
  author  = {Daw, Nathaniel D. and Niv, Yael and Dayan, Peter},
  title   = {\href{https://www.nature.com/articles/nn1560}{Uncertainty-Based Competition Between Prefrontal and Dorsolateral Striatal Systems for Behavioral Control}},
  journal = {Nature Neuroscience},
  volume  = {8},
  number  = {12},
  pages   = {1704--1711},
  year    = {2005}
}

@incollection{grice1975logic,
  author    = {Grice, H. Paul},
  title     = {\href{https://www.sfu.ca/~jeffpell/Cogs300/GriceLogicConvers75.pdf}{Logic and Conversation}},
  booktitle = {Syntax and Semantics, Volume 3: Speech Acts},
  editor    = {Cole, Peter and Morgan, Jerry L.},
  publisher = {Academic Press},
  address   = {New York},
  pages     = {41--58},
  year      = {1975}
}

@misc{zhao2025failures,
  author       = {Zhu, Kunlun and Liu, Zijia and Li, Bingxuan and others},
  title        = {\href{https://arxiv.org/abs/2509.25370}{Where {LLM} Agents Fail and How They Can Learn From Failures}},
  year         = {2025},
  howpublished = {arXiv:2509.25370},
  eprint       = {2509.25370},
  archivePrefix= {arXiv},
  primaryClass = {cs.AI}
}

@misc{arunkumar2026agentic,
  author       = {Arunkumar, V. and Gangadharan, G. R. and Buyya, Rajkumar},
  title        = {\href{https://arxiv.org/abs/2601.12560}{Agentic Artificial Intelligence ({AI}): Architectures, Taxonomies, and Evaluation of Large Language Model Agents}},
  year         = {2026},
  howpublished = {arXiv:2601.12560},
  eprint       = {2601.12560},
  archivePrefix= {arXiv},
  primaryClass = {cs.AI}
}

@misc{shen2026tripbench,
  author       = {Shen, Yuanzhe and Huang, Zisu and Wang, Zhengyuan and others},
  title        = {\href{https://arxiv.org/abs/2602.01675}{{TRIP}-Bench: A Benchmark for Long-Horizon Interactive Agents in Real-World Scenarios}},
  year         = {2026},
  howpublished = {arXiv:2602.01675},
  eprint       = {2602.01675},
  archivePrefix= {arXiv},
  primaryClass = {cs.AI}
}

@misc{wang2025hellwater,
  author       = {Wang, Andrew and Hager, Sophia and Asija, Adi and Khashabi, Daniel and Andrews, Nicholas},
  title        = {\href{https://arxiv.org/abs/2508.11027}{Hell or High Water: Evaluating Agentic Recovery from External Failures}},
  year         = {2025},
  note         = {{Conference on Language Modeling (COLM) 2025}},
  eprint       = {2508.11027},
  archivePrefix= {arXiv},
  primaryClass = {cs.CL}
}

@misc{swebenchABS2026,
  author       = {Yu, Boxi and Cao, Yang and Zhang, Yuzhong and others},
  title        = {\href{https://arxiv.org/abs/2603.00520}{{SWE-ABS}: Adversarial Benchmark Strengthening Exposes Inflated Success Rates on Test-based Benchmark}},
  year         = {2026},
  howpublished = {arXiv:2603.00520},
  eprint       = {2603.00520},
  archivePrefix= {arXiv},
  primaryClass = {cs.SE}
}

@misc{amabench2026,
  author       = {Zhao, Yujie and Yuan, Boqin and Huang, Junbo and others},
  title        = {\href{https://arxiv.org/abs/2602.22769}{{AMA-Bench}: Evaluating Long-Horizon Memory for Agentic Applications}},
  year         = {2026},
  note         = {{International Conference on Machine Learning (ICML) 2026}},
  eprint       = {2602.22769},
  archivePrefix= {arXiv},
  primaryClass = {cs.AI}
}

@misc{aegis2025,
  author       = {Song, Kevin and Jayarajan, Anand and Ding, Yaoyao and others},
  title        = {\href{https://arxiv.org/abs/2508.19504}{Aegis: Taxonomy and Optimizations for Overcoming Agent-Environment Failures in {LLM} Agents}},
  year         = {2025},
  howpublished = {arXiv:2508.19504},
  eprint       = {2508.19504},
  archivePrefix= {arXiv},
  primaryClass = {cs.AI}
}

@inproceedings{hsieh2024ruler,
  author       = {Hsieh, Cheng-Ping and Sun, Simeng and Kriman, Samuel and others},
  title     = {\href{https://arxiv.org/abs/2404.06654}{{RULER}: What's the Real Context Size of Your Long-Context Language Models?}},
  booktitle = {First Conference on Language Modeling (COLM)},
  year      = {2024}
}

@inproceedings{modarressi2025nolima,
  author       = {Modarressi, Ali and Deilamsalehy, Hanieh and Dernoncourt, Franck and others},
  title     = {\href{https://arxiv.org/abs/2502.05167}{{NoLiMa}: Long-Context Evaluation Beyond Literal Matching}},
  booktitle = {International Conference on Machine Learning (ICML)},
  year      = {2025}
}

@misc{longcontextsafety2025,
  author       = {Hadeliya, Tsimur and Jauhar, Mohammad Ali and Sakpal, Nidhi and Cruz, Diogo},
  title        = {\href{https://arxiv.org/abs/2512.02445}{When Refusals Fail: Unstable Safety Mechanisms in Long-Context {LLM} Agents}},
  year         = {2025},
  howpublished = {arXiv:2512.02445},
  eprint       = {2512.02445},
  archivePrefix= {arXiv},
  primaryClass = {cs.LG}
}

@inproceedings{lee2021ikea,
  author    = {Lee, Youngwoon and Hu, Edward S. and Lim, Joseph J.},
  title     = {\href{https://arxiv.org/abs/1911.07246}{{IKEA} Furniture Assembly Environment for Long-Horizon Complex Manipulation Tasks}},
  booktitle = {IEEE International Conference on Robotics and Automation (ICRA)},
  pages     = {6343--6349},
  year      = {2021}
}

@book{johnsonlaird1983mental,
  author    = {Johnson-Laird, Philip N.},
  title     = {\href{https://books.google.com/books/about/Mental_Models.html?id=9wiNAAAAMAAJ}{Mental Models: Towards a Cognitive Science of Language, Inference, and Consciousness}},
  publisher = {Harvard University Press},
  address   = {Cambridge, MA},
  year      = {1983}
}

@article{alchourron1985agm,
  author  = {Alchourr{\'o}n, Carlos E. and G{\"a}rdenfors, Peter and Makinson, David},
  title   = {\href{https://doi.org/10.2307/2274239}{On the Logic of Theory Change: Partial Meet Contraction and Revision Functions}},
  journal = {Journal of Symbolic Logic},
  volume  = {50},
  number  = {2},
  pages   = {510--530},
  year    = {1985}
}

@inproceedings{yehudai2025survey,
  author       = {Yehudai, Asaf and Eden, Lilach and Li, Alan and others},
  title     = {\href{https://arxiv.org/abs/2503.16416}{A Survey on Evaluation of {LLM}-based Agents}},
  booktitle = {Findings of the Association for Computational Linguistics: ACL 2026},
  year      = {2026}
}

@misc{jian2025metacognitive,
  author       = {Li, Ji-An and Xiong, Hua-Dong and Wilson, Robert C. and Mattar, Marcelo G. and Benna, Marcus K.},
  title        = {\href{https://arxiv.org/abs/2505.13763}{Language Models Are Capable of Metacognitive Monitoring and Control of Their Internal Activations}},
  year         = {2025},
  eprint       = {2505.13763},
  archivePrefix= {arXiv},
  primaryClass = {cs.AI},
  note         = {{Advances in Neural Information Processing Systems (NeurIPS) 2025}}
}

@misc{deng2025simura,
  author       = {Deng, Mingkai and Hou, Jinyu and Shen, Yilin and others},
  title        = {\href{https://arxiv.org/abs/2507.23773v1}{{SimuRA}: Towards General Goal-Oriented Agent via Simulative Reasoning Architecture with {LLM}-Based World Model}},
  year         = {2025},
  howpublished = {arXiv:2507.23773v1},
  eprint       = {2507.23773v1},
  archivePrefix= {arXiv},
  primaryClass = {cs.AI}
}

@misc{swebenchpro2025,
  author       = {Deng, Xiang and Da, Jeff and Pan, Edwin and others},
  title        = {\href{https://arxiv.org/abs/2509.16941}{{SWE-Bench Pro}: Can {AI} Agents Solve Long-Horizon Software Engineering Tasks?}},
  year         = {2025},
  howpublished = {arXiv:2509.16941},
  eprint       = {2509.16941},
  archivePrefix= {arXiv},
  primaryClass = {cs.SE}
}

@book{sutton2018reinforcement,
  author    = {Sutton, Richard S. and Barto, Andrew G.},
  title     = {\href{https://mitpress.mit.edu/9780262039246/reinforcement-learning/}{Reinforcement Learning: An Introduction}},
  edition   = {2nd},
  publisher = {MIT Press},
  address   = {Cambridge, MA},
  year      = {2018}
}

@misc{rail2026principles,
  author       = {Chiatti, Agnese and Cochez, Michael and Cornelio, Cristina and others},
  title        = {\href{https://arxiv.org/abs/2608.04285}{The {RAIL} Principles for Neurosymbolic {AI}: Reasoning, Assurances, Interfacing and Learning}},
  year         = {2026},
  howpublished = {arXiv:2608.04285},
  eprint       = {2608.04285},
  archivePrefix= {arXiv},
  primaryClass = {cs.AI}
}

\end{document}